\documentclass[%
aps, 
pre, 
twocolumn, 
nofootinbib, 
floatfix,
tightenlines, 
superscriptaddress, 
amsmath, 
amssymb, 
final, 
10pt
]{revtex4-1}

\usepackage{comment}
\usepackage{adjustbox}
\usepackage[dvipsnames]{xcolor}
\usepackage{hyperref}
\usepackage{hhline}
\usepackage{placeins}
\usepackage{graphicx}
\usepackage{dcolumn}
\usepackage{bm}
\usepackage{makecell}
\usepackage{natbib}
\usepackage{multirow}

\usepackage[ruled]{algorithm2e}

\hypersetup{
    colorlinks=true, 
    linkcolor=BrickRed, 
    citecolor=MidnightBlue, 
    filecolor=RedViolet,      
    urlcolor=MidnightBlue
    }

\usepackage{fancyhdr}

\begin{document}

\SetAlgoLined
\SetInd{0.5em}{1em}

\newcommand{\Addition}{\texttt{addition}}
\newcommand{\Parity}{\texttt{parity}}
\newcommand{\Index}{\texttt{index}} 
\newcommand{\Brackets}{\texttt{brackets}}
\newcommand{\BAPO}{\texttt{BAPO}}
\newcommand{\ROne}{\texttt{DeepSeek-R1-Distill}}
\newcommand{\B}{\,\mathrm{B}}
\newcommand{\CrI}[1]{\mathrm{CrI}_{#1\%}}
\newcommand{\elpd}[1]{\mathrm{elpd}_{\mathrm{#1}}}
\newcommand{\postE}[1]{\mathbb{E}[ #1 \mid \mathcal{D} ]}
\newcommand{\FAM}{FAM}
\newcommand{\VAM}{VAM}
\newcommand{\PEM}{PEM}
\newcommand{\PM}{PM}

\title{Scaling of Capability and Efficiency at Inference Time in Large Reasoning Models}

\author{Moritz~Laber}
\email{laber.m@northeastern.edu}
\affiliation{Northeastern University, Boston, Massachusetts, USA}
\affiliation{Complexity Science Hub Vienna, Vienna, Austria}

\author{Zohair~Shafi}
\affiliation{Northeastern University, Boston, Massachusetts, USA}

\author{Germans~Savcisens}
\affiliation{Northeastern University, Boston, Massachusetts, USA}

\author{Brennan~Klein}
\affiliation{Northeastern University, Boston, Massachusetts, USA}
\author{Matteo Chinazzi}
\affiliation{Northeastern University, Boston, Massachusetts, USA}

\author{Samuel~V.~Scarpino}
\affiliation{Northeastern University, Boston, Massachusetts, USA}
\affiliation{Santa Fe Institute, Santa Fe, New Mexico, USA}

\author{Albert-László~Barabási}
\affiliation{Northeastern University, Boston, Massachusetts, USA}

\author{Alessandro~Vespignani}
\affiliation{Northeastern University, Boston, Massachusetts, USA}

\author{Tina~Eliassi-Rad}
\affiliation{Northeastern University, Boston, Massachusetts, USA}
\affiliation{Santa Fe Institute, Santa Fe, New Mexico, USA}

\date{\today}

\begin{abstract}
Capability and efficiency are two key dimensions of reasoning in large language models (LLMs). Capability refers to the ability to solve a given problem correctly, whereas efficiency refers to the ability to do so with limited resources. When LLMs use Chain-of-Thought (CoT) reasoning to solve problems of controlled hardness, both the number of problems solved correctly and the number of tokens required to reach a correct answer depend on problem hardness and model size. However, how these factors jointly shape capability and efficiency remains poorly understood. Here, we use hierarchical Bayesian models to evaluate the capability and efficiency of LLMs from the \ROne{} model family across four classes of arithmetic and algorithmic reasoning problems. At a fixed model size, the probability of correctly solving an instance decays approximately exponentially with instance size, our proxy for problem hardness. The decay scale grows sublinearly with model size, indicating that larger models are more capable, but that capability gains diminish with scale. Output length grows as a power law with instance size, which serves as a proxy for difficulty. However, the parameters of this power law do not vary systematically with model size, suggesting that larger models do not become more efficient. Together, these findings reveal potential limitations of naive scaling as a strategy for developing more capable AI systems: capability improves with diminishing returns, while efficiency shows little to no improvement.
\end{abstract}

\maketitle
 
\section{Introduction}\label{sec:introduction}

Scaling laws are a cornerstone of modern artificial intelligence (AI). By linking invested resources (e.g., compute) to expected model performance (e.g., loss), they can inform model design and resource allocation~\cite{kaplan2020_scalinglaws, hoffmann2022_trainingcomputeoptimal, snell2024_scalingLLMtesttime, lai2025_surveyposttrainingscaling} and even help explain apparently ``emergent'' behavior in large language models (LLMs)~\cite{wei2022_emergentabilities, shaeffer2023_emergentmirage, ganguli2022_PredictabilitySurpriseLarge}.

Current models are developed through multistage training processes~\cite{guo2025_DeepSeekR1, openai2024_openaio1, kimiteam2025_Kimik15Scaling} and require substantial resources during deployment~\cite{lai2025_surveyposttrainingscaling, zhang2025_surveytesttimescaling}. Thus, scaling is relevant throughout the model life cycle. However, depending on the stage, the relevant resources, performance metrics, and functional relationships between them can differ. Although scaling laws are traditionally associated with power-law relationships, the term is also commonly used more broadly to describe resource--performance relationships regardless of their functional form. We use the term in this broader sense throughout the manuscript.

Classical neural scaling laws concern pretraining and describe how a model’s loss on held-out data decays approximately as a power law with parameter count, dataset size, and compute budget over ranges spanning several orders of magnitude~\cite{hestness2017_deeplearningscaling, kaplan2020_scalinglaws, hoffmann2022_trainingcomputeoptimal}. Subsequent theoretical work has sought to explain these empirical relationships~\cite{sharma2022_scalinglawsmanifold, bahri2024_explainingneuralscaling, bordelon2024_dynamicalmodelscaling}, while other studies have questioned their universality and precise functional form~\cite{caballero2022_brokenneuralscaling, sorscher2022_beyondneuralscaling}.

Modern models undergo extensive post-training, including supervised fine-tuning (SFT)~\cite{wei2021_FinetunedLanguageModels, ouyang2022_traininglanguagemodels, zhang2026_instructiontuning}, reinforcement learning from human feedback (RLHF)~\cite{christiano2017_RLHF, ziegler2020_finetuninghumanpreferences, lambert2026_RLHFbook}, and reinforcement learning with verifiable rewards (RLVR)~\cite{cobbe2021_trainingverifiers, lambert2025_tulu3, guo2025_DeepSeekR1, zhang2025_surveyrlreasoning}. Therefore, understanding how performance scales with the resources devoted to these stages is increasingly important~\cite{lai2025_surveyposttrainingscaling}. However, compared with pretraining, post-training exhibits less universal scaling behavior and has received less theoretical treatment. Although SFT can improve model performance across a range of metrics, no single functional form for SFT scaling is widely accepted, and empirical findings depend strongly on the experimental setup~\cite{wei2021_FinetunedLanguageModels, zhou2023_LIMAlessismore, chung2024_scalinginstructionfinetuned, zhang2024_scalingmeetsfinetuning, lin2024_selectingLLM}.

Similarly, RLHF has been shown to improve alignment with human preferences, but its scaling behavior depends on factors such as feedback quality and optimization choices~\cite{ouyang2022_traininglanguagemodels, gao2023_scalingrewardmodel, hou2024_doesRLHFscale}. Moreover, its reliance on human annotators may limit its scalability, motivating approaches such as reinforcement learning from AI feedback (RLAIF)~\cite{bai2022_constitutionalAI, lee2024_RLAIFvsRLHF}.

RLVR has been credited with driving recent advances in the reasoning abilities of LLMs and large reasoning models (LRMs)~\cite{guo2025_DeepSeekR1, openai2024_openaio1, kimiteam2025_Kimik15Scaling, zhang2025_surveyrlreasoning}, although the extent of its contribution remains debated~\cite{yue2025_doesrlreason, wang2025_dissectingreasoning}. As the RLVR compute budget increases, empirical studies have reported different scaling relationships, including diminishing or saturating gains in pass rate and power-law decreases in loss~\cite{devvrit2025_artscalingRL, cheng2026_IsoCompute, tan2026_scalingbehaviorsLLM}. Other studies suggest that data quality may matter more than quantity~\cite{wang2025_reinforcementlearningreasoning, mitsuhashi2026_quantifyingempiricalsupervision, wu2026_ReasoningMemorization}.

More recently, test-time compute (TTC) techniques have introduced another dimension of scaling during model deployment~\cite{lai2025_surveyposttrainingscaling, zhang2025_surveytesttimescaling}. Chain-of-Thought (CoT) reasoning is a prominent TTC technique in which models solve challenging problems through a sequence of intermediate steps~\cite{guo2025_DeepSeekR1, openai2024_openaio1, kimiteam2025_Kimik15Scaling}. Performance on mathematics and logic problems has been shown to decrease or vary non-monotonically as CoT length increases~\cite{feng2025_whatcharacterizeseffective, wu2025_whenmoreless, gema2025_ivnversescaling, yang2025_thinkingoptimalscaling}. This non-monotonic behavior suggests the existence of an optimal CoT length, with deviations from this optimum studied as ``underthinking'' and ``overthinking''~\cite{chen2025_donotthinkthatmuch, wang2025_thoughtsallover, opedal2025_logicprogramming, dang2025_firstimpressionproblem, su2025_underthinkingoverthinkingempirical}. Rather than allocating TTC to a single long CoT, parallel exploration distributes it across multiple shorter CoTs. Whether TTC is better allocated sequentially or in parallel remains an active topic of  debate~\cite{mirtaheri2025_LetMeThink, gu2026_understandingperformancegap}.

Unlike compute expended during pretraining, SFT, and post-training reinforcement learning, inference-time compute incurs a marginal cost for every query rather than being amortized across queries. This raises the question of how performance gains from TTC depend on model size~\cite{snell2024_scalingLLMtesttime}. More broadly, it motivates a holistic view of scaling in which resources are allocated optimally across the entire model life cycle.

Beyond scaling with resources, researchers have asked whether performance on reasoning problems exhibits a law-like dependence on problem hardness. Although performance generally declines as problem hardness increases, the functional form of this decline appears non-universal and sensitive to experimental details~\cite{mirzadeh2024_GSMSymbolic, shojaee2025_illusionthink, bie2026_AIrithmetic, zhou2025_GSMinfty, zhang2025_reasoningmeetslaws}. Efforts to characterize this dependence, coupled with the demand for minimally contaminated test sets, have spurred the development of new reasoning benchmarks and synthetic dataset generators~\cite{lin2025_zebralogic, du2025_OckBench, shi2025_KORGym, chen2025_Enigmata}.

Our main contributions characterize capability and efficiency scaling in LRMs through extensive experiments on five model sizes from the \ROne{} model family, ranging up to $70\B$ parameters and evaluated on more than $4{,}000$ reasoning problem instances per model size:

\begin{enumerate}
\item \textbf{Exponential performance decline:} We show that the probability of correctly solving a problem instance declines exponentially with instance size.
\item \textbf{Capability scaling:} We find that the characteristic scale of this exponential decay increases as a power law with the number of model parameters, indicating that larger models can solve larger and harder problem instances.
\item \textbf{Output-length scaling:} We find that, among correctly solved instances, output length increases approximately as a power law with instance size.
\item \textbf{Absence of efficiency scaling:} We observe that, for the studied model family, the prefactor and exponent governing output-length scaling vary only weakly with the number of model parameters, providing little evidence that larger models are more token efficient.
\end{enumerate}

\textbf{Paper outline.} We first introduce our evaluation strategy (Sec.~\ref{sec:results:evaluation}). Next, we examine how the number of correctly solved instances varies with instance size and model parameter count (Sec.~\ref{sec:results:capability}). We then analyze output length among correctly solved instances as a function of the same variables (Sec.~\ref{sec:results:efficiency}). Finally, we discuss the implications, limitations, and broader impact of our work (Sec.~\ref{sec:discussion}). Methodological details are provided in Sec.~\ref{sec:methods}.

\section{Results}

\subsection{Evaluation Strategy}\label{sec:results:evaluation}

\begin{figure*}
    \centering
    \includegraphics[width=0.99\textwidth]{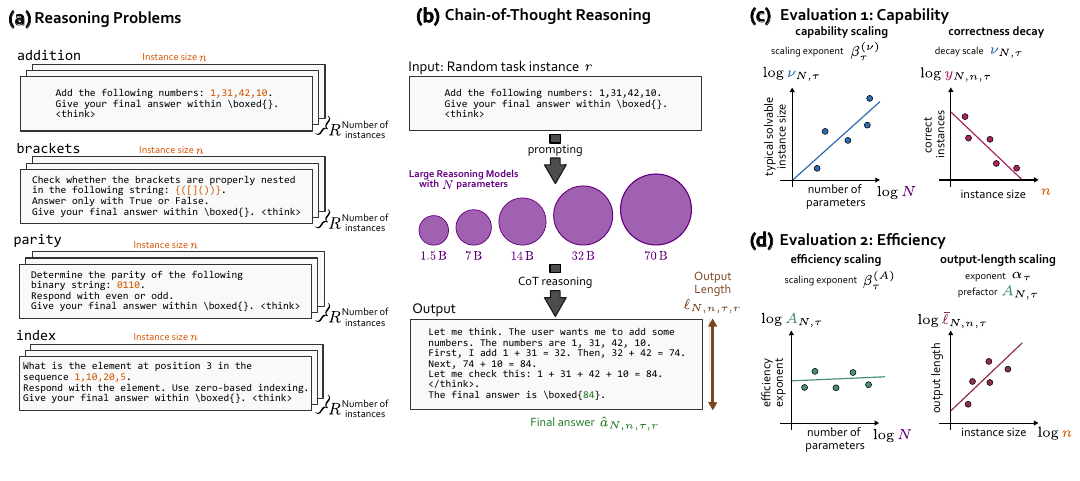}
    \caption{\textbf{Evaluation setup.} \textbf{(a)} We evaluate models from the \ROne{} family on $R$ randomly generated instances from each of four reasoning tasks: \Addition{}, \Brackets{}, \Index{}, and \Parity{} (example prompts shown). These tasks allow us to control problem hardness through the instance size $n$. \textbf{(b)} We prompt models with different parameter counts $N$ to solve these instances using CoT reasoning. For the answer $\hat{a}_{N,n,\tau,r}$ produced by a model with $N$ parameters for instance $r$ of size $n$ at temperature $\tau$, we record whether the answer is correct and measure the output length $\ell_{N,n,\tau,r}$. \textbf{(c)} To assess capability scaling, we first quantify, at fixed $N$ and $\tau$, the exponential decay in the number of correctly solved instances $y_{N,n,\tau}$ with instance size $n$. We then determine how the corresponding decay scale $\nu_{N,\tau}$ depends on model size $N$. \textbf{(d)} To assess efficiency scaling, we first model the average output length of correct responses, $\bar{\ell}_{N,n,\tau}$, as a power law in instance size $n$. We then evaluate whether the parameters of this relationship, i.e., its prefactor $A_{N,\tau}$ and exponent $\alpha_{N,\tau}$, scale with model size $N$.}
    \label{fig:schematic}
\end{figure*}

Our evaluation strategy treats capability and efficiency as latent model traits that cannot be observed directly. We infer them from the number of correctly solved instances and the observed output lengths, respectively. Similarly, how these traits scale with model size $N$ is a latent property of the model family that must be inferred. Figure~\ref{fig:schematic} provides a schematic overview of our approach.

In our experiments, we evaluate five LRMs from the \ROne{} family~\cite{guo2025_DeepSeekR1}, with parameter counts $N\in \{1.5\B, 7\B, 14\B, 32\B, 70\B\}$. We sample responses at temperatures $\tau \in \{0.4, 0.6, 0.8\}$. Because $\tau=0.6$ is the recommended setting, we use it as our primary setting and use the other temperatures to assess the robustness of our findings to changes in sampling temperature. Further details are provided in Methods (Sec.~\ref{sec:methods:LRMs}).

We evaluate these models on randomly generated instances from four reasoning tasks with tunable hardness: \Addition{}, \Brackets{}, \Parity{}, and \Index{}. In the \Addition{} task, the model must sum $n$ integers drawn from the set $\{0,\dots,100\}$. The \Brackets{} task asks the model to determine whether a sequence of $n$ bracket symbols is properly nested. In the \Parity{} task, the model must determine whether the number of ones in a length-$n$ binary string is even or odd. Finally, an instance of the \Index{} task consists of a list of $n$ integers drawn from $\{0,\dots,100\}$ and a position $0\leq i <n$. The model must return the element at that position. We formally define these tasks and describe how instances are generated in Methods (Sec.~\ref{sec:methods:tasks}). Table~\ref{tab:tasks} summarizes the tasks, their hardness parameters, and example prompts.

These tasks have several desirable properties. First, their difficulty can be controlled through the instance size $n$, with larger instances generally being more challenging than smaller ones. For example, summing $10$ numbers is reasonably assumed to be less challenging than summing $100$ numbers. Second, all four tasks admit polynomial-time algorithms. Thus, in principle, models can solve every instance using an efficient procedure rather than relying on guesses, heuristics, or exhaustive search over possible solutions. These algorithms also allow us to verify model answers efficiently. Third, task hardness can be characterized not only through classical measures of time and space complexity but also through transformer-specific frameworks such as Bounded Attention Prefix Oracle (\BAPO{}) complexity~\cite{schnabel2025_losttransmission,tomlinson2026_ReasoningReasoningBAPO}. Table~\ref{tab:tasks} lists the time, space, and \BAPO{} complexity of each task and shows that our evaluation covers tasks that are \BAPO{}-easy and tasks that are conjectured to be \BAPO{}-hard. We discuss task hardness in greater detail in Methods (Sec.~\ref{sec:methods:tasks}).

For each task and instance size $n$, we generate $R=100$ independent instances and determine the correct answer $a_{n,r}$ for each instance $r$. We use instance sizes $n\in \{2,10,50, 100,150,\dots,500\}$. We then prompt each model with $N$ parameters to solve each instance at temperature $\tau$ and extract the predicted answer $\hat{a}_{N,n,\tau,r}$ from its output. We also record the output length in tokens of the model's native tokenizer, denoted by $\ell_{N,n,\tau,r}$. We describe the answer-extraction procedure in Methods (Sec.~\ref{sec:methods:answer-extraction}).

The two key observables of our study are the number of correctly solved instances and the average output length among correct responses. To determine them, we define the set $\mathcal{C}_{N,n,\tau}$ of instances of size $n$ correctly answered by a model of size $N$ at temperature $\tau$,
\begin{equation}\label{eq:correct_set}
    \mathcal{C}_{N,n,\tau}
    =
    \left\{
        r \in \{1,\dots,R\}
        :
        \hat{a}_{N,n,\tau,r} = a_{n,r}
    \right\}
    .
\end{equation}
The number of instances of size $n$ that are correctly solved by a model of size $N$ at temperature $\tau$ is
\begin{equation}\label{eq:num_correct}
    y_{N,n,\tau}
    =
    \left| \mathcal{C}_{N,n,\tau} \right|
    .
\end{equation}
The average output length among correct responses from a model of size $N$ at temperature $\tau$ to an instance of size $n$ is
\begin{equation}\label{eq:avg_len_correct}
    \bar{\ell}_{N,n,\tau}
    =
    \frac{1}{y_{N,n,\tau}}
    \sum_{r \in \mathcal{C}_{N,n,\tau}}
    \ell_{N,n,\tau,r}
    ,
\end{equation}
and its variance is $s_{N,n,\tau}^2$
\begin{equation}\label{eq:var_len_correct}
    s^2_{N,n,\tau}
    =
    \frac{1}{y_{N,n,\tau}}
    \sum_{r \in \mathcal{C}_{N,n,\tau}}
    (
        \ell_{N,n,\tau,r} - \bar{\ell}_{N,n,\tau}
    )^2. 
\end{equation}
Equations~\eqref{eq:avg_len_correct} and~\eqref{eq:var_len_correct} are defined only when the model solves at least one instance correctly, that is, when $y_{N,n,\tau}>0$. We exclude configurations that do not satisfy this condition from our analysis of model efficiency.

To infer model capability and its dependence on model size $N$, we use hierarchical Bayesian models. To distinguish them from the reasoning models under study, we refer to them as \textit{statistical models} throughout.

For an LRM with $N$ parameters evaluated at temperature $\tau$, we model the number $y_{N,n,\tau}$ of correctly solved instances among $R$ independently generated size-$n$ problem instances as
\begin{equation}\label{eq:num-correctly-solved}
y_{N,n,\tau}\sim\operatorname{Binomial}\left(R,p_{N,\tau}(n)\right),
\end{equation}
where $p_{N,\tau}(n)$ is the probability of correctly solving a size-$n$ instance. We assume $p_{N,\tau}(n)$ decays exponentially in $n$ with decay constant $\nu_{N,\tau}$,
\begin{equation}\label{eq:exponential-decay-p}
    p_{N,\tau}(n)
    =
    p_{N, \tau}^{(\infty)}
    +
    \left(
        p_{\tau}^{(0)} - p_{N,\tau}^{(\infty)}
    \right)
    \exp
    \left(
        -\frac{n}{\nu_{N,\tau}}
    \right)
    ,
\end{equation}
where $p_{\tau}^{(0)}\in [0,1]$ governs the probability that a hypothetical size $n = 0$ instance is solved correctly, and the large-$n$ asymptote $p_{N,\tau}^{(\infty)}\in [0,1]$ is the limiting probability that an instance is solved correctly as $n\to\infty$.\footnote{We require $p_{N,\tau}^{(\infty)}\leq p_{\tau}^{(0)}$ to ensure that performance declines with instance size.} The parameter $\nu_{N,\tau}>0$ has the same units as $n$ and determines the characteristic scale over which performance decays. We treat $\nu_{N,\tau}$ as a measure of the capability of a model with $N$ parameters evaluated at temperature $\tau$.

To understand how capability scales with model size $N$, we assume that the capabilities within a model family are connected through a latent scaling law, i.e.,
\begin{equation}\label{eq:capability-scaling}
    \log \nu_{N,\tau}
    =
    \log C_{\tau}^{(\nu)}
    +
    \beta_{\tau}^{(\nu)}
    \log \frac{N}{\tilde{N}}
    +
    \eta_{\tau}^{(\nu)} \xi_{N,\tau}^{(\nu)},
\end{equation}
where $\beta_{\tau}^{(\nu)}$ is the scaling exponent, $C_{\tau}^{(\nu)}$ is the prefactor, and $\eta_{\tau}^{(\nu)}$ determines the standard deviation of the multiplicative Gaussian noise $\xi_{N,\tau}^{(\nu)}\sim\mathcal{N}(0,1)$ that accounts for deviations from the scaling law, e.g., through finite-size effects. We use $\nu_\tau(N)$ to denote the quantity in Eq.~\eqref{eq:capability-scaling} without these deviations. We rescale all model sizes $N$ under consideration by their geometric mean $\tilde{N}$ for numerical stability. 

We study two versions of this statistical model. In the version defined above, which we call the \textit{variable asymptote model} (\VAM{}), the large-$n$ asymptote $p_{N,\tau}^{(\infty)}$ may vary with model size $N$. In the alternative \textit{fixed asymptote model} (\FAM{}), the asymptote is shared across model sizes, such that $p_{N,\tau}^{(\infty)}=p_{\tau}^{(\infty)}$.

To fully specify these hierarchical Bayesian models, we need to put priors on all model parameters. We discuss the choice of priors in Methods Sec.~\ref{sec:methods:statistical-models}. We infer all parameters using Monte Carlo sampling and provide details on sampling hyperparameters in Methods Sec.~\ref{sec:methods:MCMC}.

We take a similar approach to modeling efficiency, operationalized as using few tokens for solving a task, but adjust our parametric assumptions accordingly. We assume that the logarithm of the average output length among correct responses produced by a model with $N$ parameters at temperature $\tau$ follows a normal distribution with mean
\begin{equation}\label{eq:avg_len_scaling_mean}
    \log \lambda_{N,n,\tau} 
    =
    \log A_{N,\tau}
    +
    \alpha_{N,\tau}
    \log \frac{n}{\tilde{n}},
\end{equation}
where $\tilde{n}$ is the geometric mean of the considered instance sizes.
The corresponding variance is
\begin{equation}\label{eq:avg_len_scaling_var}
    \omega_{N,n,\tau}^2
    =
    \frac{
        s^2_{N,n,\tau
    }}{
        y_{N,n,\tau} \lambda_{N,n,\tau}^2
    }
    +
    \zeta_{\tau}^2
    ,
\end{equation}
where the first term captures the sampling variance of the estimated mean at fixed model size $N$, temperature $\tau$, and instance size $n$, while the second term, $\zeta_\tau^2$, captures excess variation. We derive the form of $\omega_{N,n,\tau}^2$ in Appendix~\ref{app:delta-method}.

We compare two statistical models of scaling across model sizes. The more flexible statistical model, which we call the \textit{prefactor exponent model} (\PEM{}), accounts for the possibility that both the prefactor $A_{N,\tau}$ and the scaling exponent $\alpha_{N,\tau}$ in Eq.~\eqref{eq:avg_len_scaling_mean} scale with the reasoning model size $N$, i.e.,
\begin{equation}\label{eq:len_prefactor_scaling}
    \log A_{N,\tau}
    =
    \log C^{(A)}_{\tau}
    +
    \beta^{(A)}_{\tau}
    \log \frac{N}{\tilde{N}}
    +
    \eta_{\tau}^{(A)} \xi_{N,\tau}^{(A)}
\end{equation}
and 
\begin{equation}\label{eq:len_exponent_scaling}
    \log \alpha_{N,\tau}
    =
    \log C^{(\alpha)}_{\tau}
    +
    \beta^{(\alpha)}_{\tau}
    \log \frac{N}{\tilde{N}}
    +
    \eta_{\tau}^{(\alpha)}\xi_{N, \tau}^{(\alpha)},
\end{equation}
where $C^{(A)}_{\tau}$, $C^{(\alpha)}_\tau$ are prefactors, $\beta_{\tau}^{(A)}$, $\beta_{\tau}^{(\alpha)}$ are scaling exponents, and $\eta_{\tau}^{(A)}$, $\eta_{\tau}^{(\alpha)}$ govern the magnitudes of the multiplicative Gaussian noise $\xi_{N,\tau}^{(A)}$ and $\xi_{N,\tau}^{(\alpha)}$, respectively.  We use $A_{\tau}(N)$ and $\alpha_{\tau}(N)$ to denote the quantities in Eqs.~\eqref{eq:len_prefactor_scaling} and~\eqref{eq:len_exponent_scaling}, respectively, with the random effects set to zero.

A parsimonious alternative statistical model is the \textit{prefactor model} (\PM{}), in which the exponent $\alpha_{N,\tau}$ has no systematic dependence on $N$. The \PM{} can be obtained from the \PEM{} by fixing $\beta_{\tau}^{(\alpha)}=0$.

When presenting our results, we use $\postE{\theta}$ to denote the posterior mean of a parameter $\theta$ conditional on the relevant data $\mathcal{D}$. We also use the notation $\mathrm{CrI}_{95\%}[\theta]$ to denote the $95\%$ credible interval of $\theta$.

\subsection{Capability}\label{sec:results:capability}

\begin{figure*}[tb]
    \centering
    \includegraphics[width=\linewidth]{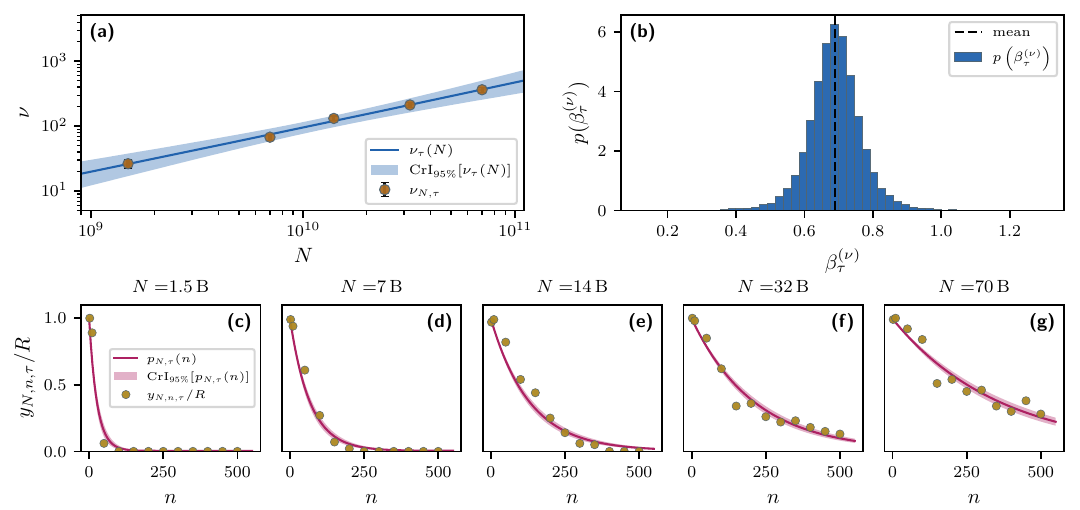}
    \caption{\textbf{Capability scaling.} Fitting the \FAM{} to data from the \Addition{} task at temperature $\tau=0.6$ suggests that \textbf{(a)} the exponential decay scale $\nu_{N,\tau}$ (orange circles) increases approximately as a power law with $N$, and deviations from the latent scaling curve $\nu_\tau(N)$ (blue line) are small. \textbf{(b)} The posterior of the scaling exponent concentrates on positive values smaller than one and has mean $\postE{\beta_{\tau}^{(\nu)}}\approx 0.69$ (black dashed line). \textbf{(c)--(g)} The probability of correctly solving an instance, $p_{N,\tau}(n)$ (magenta lines), decays exponentially toward a shared asymptote and closely follows the observed fraction of correctly solved instances $y_{N,n,\tau}/R$ (yellow circles) as a function of instance size $n$ for increasing model sizes $N$ (at \textbf{(c)} $N=1.5\B$, \textbf{(d)} $N=7\B$, \textbf{(e)} $N=14\B$, \textbf{(f)} $N=32\B$, and \textbf{(g)} $N=70\B$). Thus, for models in the \ROne{} family, the probability of correctly solving an instance decays exponentially with instance size $n$, while capability, measured by the decay scale $\nu_\tau(N)$, grows sublinearly with model size $N$.
    }
    \label{fig:capability-FAM-addition-0.6}
\end{figure*}

We illustrate our approach to quantifying the capability of an LRM using the \Addition{} task at temperature $\tau=0.6$ under the \FAM{}. We provide detailed results for the other tasks (\Brackets{}, \Index{}, and \Parity{}), temperatures  ($\tau\in \{0.4,0.8\}$), and the \VAM{} in Appendix~\ref{app:additional-results:capability}.

Fitting the \FAM{} to the results of the \Addition{} task at $\tau=0.6$ provides evidence that the characteristic instance-size scale $\nu_{N,\tau}$ scales with the number of parameters $N$ (Fig.~\ref{fig:capability-FAM-addition-0.6}\textbf{(a)}). The inferred scale $\eta_\tau^{(\nu)}$ of deviations from the latent scaling law $\nu_\tau(N)$ is small, with a posterior mean of $\postE{\eta^{(\nu)}_{\tau}}\approx0.20$.

The inferred scaling is sublinear, with a posterior mean of the scaling exponent $\postE{\beta^{(\nu)}_{\tau}} \approx 0.69$ and a $95\%$ credible interval of $\CrI{95}[\beta^{(\nu)}_{\tau}]\approx (0.53, 0.85)$ (Fig.~\ref{fig:capability-FAM-addition-0.6}\textbf{(b)}). At the posterior mean, doubling the parameter count of a \ROne{} model increases the characteristic instance-size scale of the exponential decline by a factor of $2^{0.69} \approx 1.61$.

Inspecting the performance decay at fixed parameter counts (Fig.~\ref{fig:capability-FAM-addition-0.6}\textbf{(c)}--\textbf{(g)}), we find that the exponential decay with a shared asymptote, as assumed in the \FAM{}, describes the data well. The probability of correctly solving an instance, $p_{N,\tau}(n)$, recapitulates the trends in the observed fraction of correctly solved instances $y_{N,n,\tau}/R$. Moreover, the \FAM{} has a higher expected log predictive density ($\elpd{\FAM{}}\approx-199$) than the \VAM{} ($\elpd{\VAM{}}\approx-212$) and is therefore favored by this criterion.  Our fits suggest that the scale of the exponential decay increases from $\postE{\nu_{N,\tau}} \approx 26.4$ for the smallest LRM, with $N=1.5\B$ parameters, to $\postE{\nu_{N,\tau}} \approx 363.5$ for the largest LRM, with $N=70\B$ parameters. This provides a rough estimate of the size of an \Addition{} instance that an LRM can typically solve.

\begin{figure*}
    \centering
    \includegraphics[width=\linewidth]{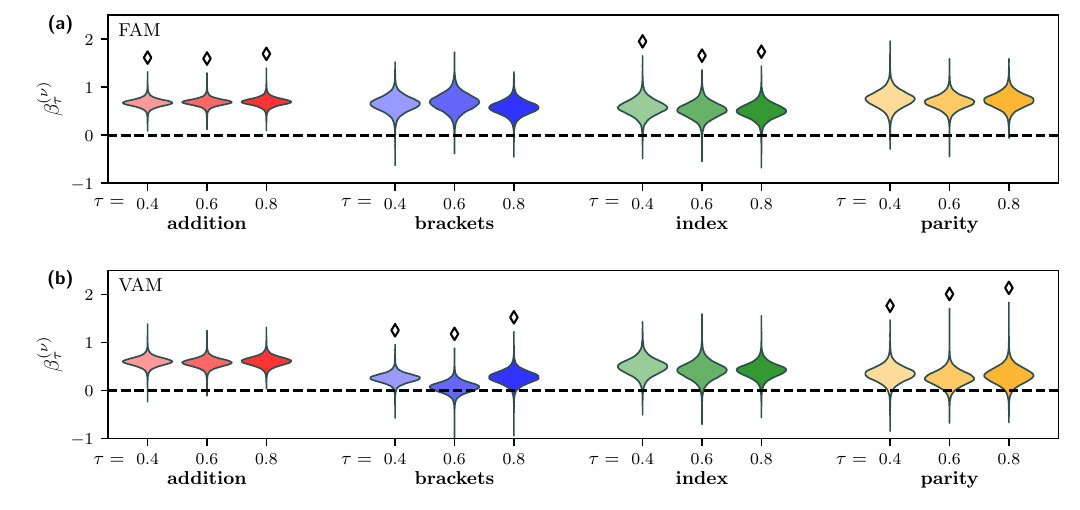}
    \caption{\textbf{Capability comparison.} Comparing the posterior distributions of the scaling exponent $\beta_{\tau}^{(\nu)}$ across statistical models (\textbf{(a)} \FAM{}, \textbf{(b)} \VAM{}), tasks (\Addition{} (red), \Brackets{} (blue), \Index{} (green), and \Parity{} (yellow)), and temperatures (shades), we find that all posterior distributions place most of their mass on positive values of $\beta_{\tau}^{(\nu)}$ lower than one and therefore favor sublinear scaling across experimental conditions and statistical models. Note that the statistical model with the higher estimated $\elpd{}$ varies across conditions, as indicated by $\boldsymbol{\diamond}$.}
    \label{fig:capability-comparison}
\end{figure*}

Across tasks and temperatures, we find that the posterior means of the scaling exponent $\beta^{(\nu)}_\tau$ indicate a sublinear scaling of the performance decay parameter $\nu_\tau(N)$ with the number of LRM parameters $N$, irrespective of whether the \FAM{} (Fig.~\ref{fig:capability-comparison}\textbf{(a)}) or the \VAM{} (Fig.~\ref{fig:capability-comparison}\textbf{(b)}) is used. We indicate which model has the higher estimated $\elpd{}$ using the $\boldsymbol{\diamond}$ symbol and provide detailed model-comparison results in Appendix~\ref{app:model-comparison}.

For the \Addition{} task, the posterior means of the scaling exponent $\beta_\tau^{(\nu)}$ vary little across temperatures, and the corresponding posterior  distributions are narrowly concentrated. Although the \FAM{} has a higher $\elpd{}$ than the \VAM{} across temperatures, our qualitative conclusions are unchanged by the choice of statistical model.

For the \Brackets{} task, the posterior means $\postE{\beta^{(\nu)}_{\tau}}$ are higher under the \FAM{} than under the \VAM{} across temperatures $\tau$. Nevertheless, both statistical models are compatible with the capability scale increasing sublinearly with $N$. The \VAM{} has a higher estimated $\elpd{}$ than the \FAM{} because it accommodates model-size-dependent asymptotes. For larger LRMs, the inferred large-$n$ asymptote $Rp_{N,\tau}^{(\infty)}$ is close to the random-guessing expectation of $\frac{R}{2}=50$ correctly solved instances, whereas for smaller LRMs, it is close to zero because they often fail to provide an answer (see Appendix~\ref{app:additional-results:capability}).

For the \Index{} task, both statistical models yield similar posterior means $\postE{\beta_\tau^{(\nu)}}$ across temperatures. These posterior means lie between $0$ and $1$, indicating sublinear scaling of the characteristic decay scale $\nu_\tau(N)$ with the LRM parameter count $N$. The \FAM{} has a slightly higher estimated $\elpd{}$ than the \VAM{}.

For the \Parity{} task, the more flexible \VAM{} has a higher estimated $\elpd{}$ than the \FAM{}. Under both statistical models, the posterior means $\postE{\beta_\tau^{(\nu)}}$ lie between $0$ and $1$, indicating sublinear scaling, but they are lower under the \VAM{} than under the \FAM{}. Within each statistical model, the posterior means vary little across temperatures $\tau$.

Together, these findings suggest that larger models in the \ROne{} family are able to solve larger instances of our four reasoning tasks. Across all experiments, the posterior means of the capability-scaling exponent satisfy $0<\postE{\beta_\tau^{(\nu)}}<1$, indicating sublinear growth of the  decay scale $\nu_\tau(N)$ with model size $N$ at the level of the posterior means.

\subsection{Efficiency}\label{sec:results:efficiency}

\begin{figure*}[tb]
    \centering
    \includegraphics[width=\linewidth]{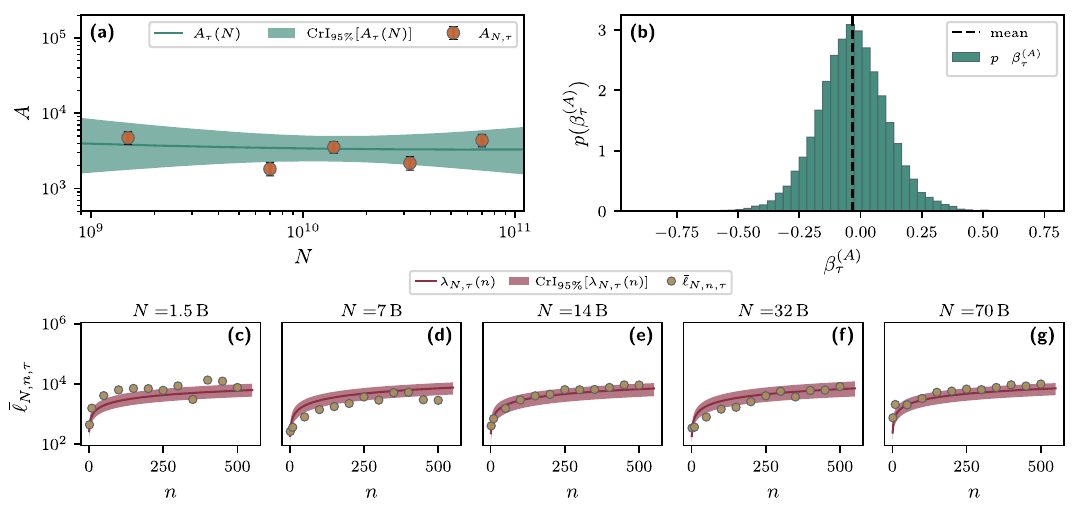}
    \caption{\textbf{Efficiency scaling.} Fitting the \PM{} to data from the \Index{} task at temperature $\tau = 0.6$ suggests that \textbf{(a)} the prefactor $A_{N,\tau}$ of the length scaling (orange circles) approximately follows an almost flat latent scaling law $A_{\tau}(N)$ (teal line). \textbf{(b)} The posterior of the latent scaling exponent $\beta_{\tau}^{(A)}$ is concentrated around zero, $\postE{\beta_{\tau}^{(A)}}\approx -0.03$ (black dashed line). A scaling ansatz with prefactors $A_{N,\tau}$ that vary systematically with model size $N$ and exponents $\alpha_{N,\tau}$ that vary only through random effects (red line) offers a good fit to the observed output lengths $\bar{\ell}_{N,n,\tau}$ (yellow circles) as a function of the instance size $n$ across model sizes $N$ (\textbf{(c)} $N=1.5\B$, \textbf{(d)} $N=7\B$, \textbf{(e)} $N=14\B$, \textbf{(f)} $N=32\B$, \textbf{(g)} $N=70\B$). These results suggest that while output length approximately increases as a power law in the instance size, the model efficiency, measured by the prefactor of this power law, does not scale substantially with model size $N$.}
    \label{fig:efficiency-PM-index-0.6}
\end{figure*}

Figure~\ref{fig:efficiency-PM-index-0.6} illustrates how we evaluate the scaling of model efficiency with the number of parameters $N$ using the \PM{} for the $\Index{}$ task and temperature $\tau = 0.6$. Results for the remaining temperatures ($\tau \in \{0.4, 0.8\}$) and other tasks can be found in Appendix~\ref{app:additional-results:efficiency}.

For the \Index{} task, we find that the inferred scaling of model efficiency, quantified by the prefactor $A_{N,\tau}$ in Eq.~\eqref{eq:avg_len_scaling_mean}, with model parameter count $N$ is approximately flat (Fig.~\ref{fig:efficiency-PM-index-0.6}\textbf{(a)}).
The model attributes deviations from the largely flat latent scaling curve $A_{\tau}(N)$ to a log-scale random effect with posterior mean standard deviation $\postE{\eta^{(A)}_\tau} \approx 0.42$. The posterior mean of the scaling exponent $\beta_{\tau}^{(A)}$ is $\postE{\beta_{\tau}^{(A)}} \approx -0.03$, and its $95\%$ credible interval straddles zero, $\CrI{95}[\beta_{\tau}^{(A)}]\approx (-0.32, 0.25)$ (Fig.~\ref{fig:efficiency-PM-index-0.6}\textbf{(b)}).

At fixed model sizes $N$ on the \Index{} task (Fig.~\ref{fig:efficiency-PM-index-0.6}\textbf{(c)}--\textbf{(g)}), the \PM{} captures the increase in average output length $\bar{\ell}_{N,n,\tau}$ with instance size $n$. The inferred scaling is sublinear, with posterior means $\postE{\alpha_{N,\tau}}$ ranging from $0.52$ to $0.57$. This implies that doubling the instance size increases average output length by a factor between $2^{0.52}\approx1.43$ and $2^{0.57}\approx1.48$.

\begin{figure*}
    \centering
    \includegraphics[width=\linewidth]{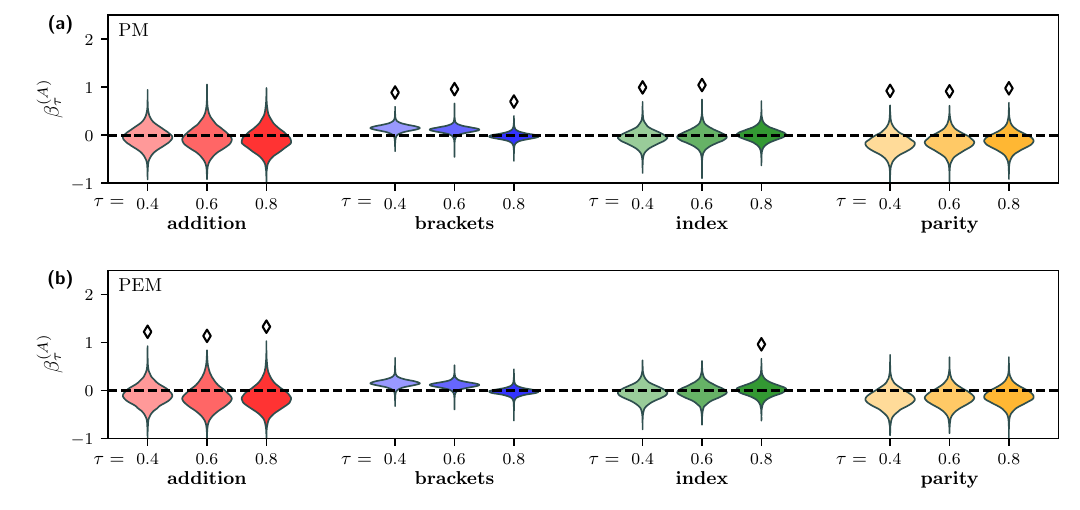}
    \caption{\textbf{Efficiency comparison.} Comparing the posterior distributions of the scaling exponent $\beta_{\tau}^{(A)}$ across statistical models (\textbf{(a)} \PM{}, \textbf{(b)} \PEM{}), tasks (\Addition{} (red), \Brackets{} (blue), \Index{} (green), \Parity{} (yellow)), and temperatures (shades), provides no strong evidence for scaling of efficiency with the number of model parameters $N$. The posterior distributions are centered close to zero in most cases. Our findings are consistent across statistical models, even if different combinations of task and temperature favor different models in terms of $\elpd{}$, as indicated by $\boldsymbol{\diamond}$.}
    \label{fig:length-comparison}
\end{figure*}

Across tasks, statistical models, and temperatures, we do not find strong evidence that the prefactor $A_{N,\tau}$ in Eq.~\eqref{eq:avg_len_scaling_mean} decreases systematically with model size, as would be expected if larger models were more efficient (Fig.~\ref{fig:length-comparison}). We indicate the statistical model with the higher estimated $\mathrm{elpd}$ using the $\boldsymbol{\diamond}$ symbol and provide further details on model comparison in Appendix~\ref{app:model-comparison}.

For the \Addition{} task (Fig.~\ref{fig:length-comparison}), both the \PM{} and the \PEM{} yield posterior distributions of $\beta_{\tau}^{(A)}$ that are centered around zero. The \PEM{} achieves higher $\elpd{}$ at all temperatures and is thus favored in model comparison.

For the \Brackets{} task (Fig.~\ref{fig:length-comparison}), we find that results differ by temperature $\tau$ but are consistent between statistical models. At temperatures $\tau \in \{0.4, 0.6\}$, both the \PM{} and the \PEM{} put most of the posterior mass on small positive values of $\beta_{\tau}^{(A)}$, indicating that models at these temperatures become less efficient as model size increases, but this effect is weak. At temperature $\tau=0.8$, in contrast, the posterior covers zero. For the \Brackets{} task, the \PM{} is preferred in model comparison.

For the \Index{} task (Fig.~\ref{fig:length-comparison}), the \PM{} offers a better fit than the \PEM{} at temperatures $\tau \in \{0.4, 0.6\}$ but not at temperature $\tau = 0.8$. Both statistical models provide no strong evidence that efficiency scales with model size.  

For the \Parity{} task (Fig.~\ref{fig:length-comparison}), although the posterior distribution of $\beta_{\tau}^{(A)}$ covers zero under both statistical models, more posterior mass lies on small negative values. This is compatible with weak efficiency gains as the number of parameters $N$ increases. Model comparison favors the \PM{} over the \PEM{} at all temperatures.

Together, these results paint a nuanced picture of efficiency scaling in LRMs from the \ROne{} family. For most task and temperature combinations, efficiency does not increase with model size. In some cases, increases or decreases in efficiency might be present, but any such effect is weak.

\section{Discussion}\label{sec:discussion}

In this paper, we asked whether larger models become more capable and more efficient at solving arithmetic and algorithmic reasoning problems. We find that models become more capable, i.e., larger models can solve harder problems. However, we do not find evidence that they are more efficient, i.e., correct solutions produced by larger models are not systematically shorter than those produced by smaller models.

These findings have implications for model design, as they call into question the na\"ive scaling approach to artificial intelligence. While we find that larger models are more capable, the scaling laws we infer are sublinear. This means that scaling the number of parameters comes with diminishing returns. Every additional parameter translates into a smaller increase in the decay scale $\nu_{N,\tau}$ than the previous parameter added. One potential explanation is that model resources that are relevant for integrating information across larger problem instances, e.g., a model's width or depth, grow sublinearly in the number of parameters. Another possible explanation is that solving large problem instances draws on multiple abilities of the model, e.g., the ability to generalize across instance sizes or to produce CoTs longer than those present in the training data. Capability scaling could then be effectively limited by the ability that increases most slowly. 

Moreover, we find no strong evidence that models become more efficient. This means that upfront investment in parameter budget during pretraining does not translate into systematically shorter answers. This finding could impact how quickly pretraining cost can be amortized. One potential explanation for this observation is that algorithmically solving a problem requires executing a given number of operations, as specified by its time complexity. This fundamentally limits possible efficiency gains. Another explanation could be that token efficiency, in contrast to correctness, is not explicitly rewarded under current post-training paradigms.

We also contribute to the ongoing debate on what measuring model capability and efficiency should mean. In this work, we view both capability and efficiency as latent properties of a model. This means that these properties are not directly observable and must be inferred from behavioral data. 

Our work is not without limitations, and addressing them directly points to opportunities for future work. First, studying well our findings generalize beyond the \ROne{} model family and our four reasoning tasks is an important open question. This could also address potential confounding introduced by distilling the \texttt{DeepSeek-R1} architecture into different base models with different tokenizers. Second, our evaluation assesses capability and efficiency separately. Studying their interplay and potential latent couplings between them could uncover capability-efficiency trade-offs, elucidate the role of correctness-conditioning on efficiency, and, in the long run, aid the design of Pareto-optimal models. Finally, our statistical models are agnostic to the origin of capability and efficiency. Designing statistical procedures that decompose capability and efficiency gains along the model life cycle could guide resource-optimal training and deployment.

\section{Methods}\label{sec:methods}

\subsection{Reasoning Models and Sampling}\label{sec:methods:LRMs}

We use the following models from the \ROne{} model family~\cite{guo2025_DeepSeekR1}:
\begin{itemize}
    \setlength{\itemsep}{1pt}
    \setlength{\parskip}{1pt}
    \setlength{\parsep}{1pt}
    \item \ROne{}\texttt{-Qwen-1.5B}
    \item \ROne{}\texttt{-Qwen-7B}
    \item \ROne{}\texttt{-Qwen-14B}
    \item \ROne{}\texttt{-Qwen-32B}
    \item \ROne{}\texttt{-Llama-70B}
\end{itemize}

These models range from $N=1.5\B$ to $N=70\B$ parameters and are obtained via distillation of the full \texttt{DeepSeek-R1} model with $N=671\B$ parameters into smaller Qwen or Llama models. We use temperature sampling with temperature $\tau \in \{0.4, 0.6, 0.8\}$ in all our experiments with $\tau = 0.6$ being the recommended temperature for deployment. If the end-of-sequence token is not emitted after $75{,}000$ sampled tokens, we stop the generation process.

We leverage the \texttt{SGLang} library~\cite{zheng2024_sglang} (Apache 2.0 License) for continuous batching as well as parallelization across NVIDIA H200 and H100 GPUs.

\subsection{Reasoning Tasks}\label{sec:methods:tasks}

\begin{table*}[tb]
    \centering
    \begin{tabular}{|l|c|c|c|c|}
    \hline
    Task              &  \BAPO{} complexity & Time complexity  & Space complexity & Example Prompt \\ \hline
    \Addition{}        &  \makecell[c]{\BAPO{}-hard \\ (conjectured)}         & $\Theta(n)$ &  $\mathcal{O}(1)$      & \makecell[c]{\texttt{Add the following numbers:}\\ \texttt{1, 21, 40, 12}\texttt{.\textbackslash{}n Give your final} \\ \texttt{answer within \textbackslash boxed\{\}\textbackslash{}n <think>}} \\ [15pt]
    \Brackets{}         & \makecell[c]{\BAPO{}-hard \\ (conjectured)}        & $\Theta(n)$ & $\mathcal{O}(n)$      & \makecell[c]{\texttt{Check whether the brackets are properly} \\ \texttt{closed and nested in the following} \\ \texttt{string:} \texttt{(()\{[]\}[])}\texttt{.\textbackslash{}n Answer} \\ \texttt{only with True or False. Give your} \\ \texttt{final answer within \textbackslash{}boxed\{\}.\textbackslash{}n <think>}} \\ [25pt]
    \Parity{}           & \BAPO{}-easy         & $\Theta(n)$ & $\mathcal{O}(1)$     & \makecell[c]{\texttt{Determine the parity of the following} \\ \texttt{binary string:} \texttt{1010110}\texttt{.\textbackslash{}n Respond with even} \\ \texttt{or odd. Give your final answer within} \\ \texttt{\textbackslash{}boxed\{\}.\textbackslash{}n <think>}} \\ [20pt]
    \Index{}            & \BAPO{}-easy         & $\mathcal{O}(1)$ to $\mathcal{O}(n)$ & $\mathcal{O}(1)$      & \makecell[c]{\texttt{What is the element at position} $3$ \\ \texttt{in the sequence} \texttt{10, 2, 11, 20}\texttt{?} \texttt{\textbackslash{}n Respond} \\ \texttt{with the element. Use zero-based indexing.} \\ \texttt{Give your final answer within \textbackslash{}boxed\{\}.} \\\texttt{\textbackslash{}n <think>}}\\
    \hline
    \end{tabular}
    \caption{\textbf{Reasoning tasks.} Overview of our four reasoning tasks (\Addition{}, \Brackets{}, \Parity{}, and \Index{}) including their conjectured \BAPO{} complexity, time complexity, and space complexity, as well as an example prompt for each task.}
    \label{tab:tasks}
\end{table*}

Here, we describe the four reasoning tasks in more detail, including how random instances are constructed and how the correct answer is computed. We also comment on the time, space, and \BAPO{} complexity of each task. An overview can be found in Tab.~\ref{tab:tasks}.

\subsubsection{Addition}

A size-$n$ instance of the \Addition{} task consists of a multiset of integers $\mathcal{X}$ of size $n$. In our experiments, we construct random instances of the addition task by drawing the elements of $\mathcal{X}$ uniformly at random with replacement from $\{0,1,\dots, 100\}$. 

The correct answer $a$ is computed by summing all elements of $\mathcal{X}$,
\begin{align}
    a
    =
    \sum_{x \in \mathcal{X}}
    x
    .
\end{align}

Assuming that two integers can be added in $\mathcal{O}(1)$ time, and that storing an integer requires $\mathcal{O}(1)$ space, an algorithm that maintains the running sum $S$ and adds numbers one by one has runtime $\Theta(n)$ and requires $\mathcal{O}(1)$ space for the variable $S$. As adding $n$ numbers requires touching each number at least once, there is no faster sequential algorithm.

Neither Schnabel et al.~\cite{schnabel2025_losttransmission} nor Tomlinson et al.~\cite{tomlinson2026_ReasoningReasoningBAPO} establish \BAPO{} complexity for \Addition{}. However, they establish that the \texttt{majority} task, which consists of deciding whether the number of ``$1$'' symbols in a binary string is larger than the number of ``$0$'' symbols, is \BAPO{}-hard. We conjecture that a constant bandwidth \BAPO{} for solving \Addition{} of integers from $\{0,1\}$ could be used to solve \texttt{majority}. If this conjecture is true, no such \BAPO{} can exist without contradicting Theorem 4 of Schnabel et al.~\cite{schnabel2025_losttransmission}. Thus, provided the conjecture holds, \Addition{} is \BAPO{}-hard.

\subsubsection{Brackets}

To solve a size-$n$ instance of the \Brackets{} task, the model needs to decide whether a sequence $\mathbf{x}=(x_0,\dots,x_{n-1})$ of $n$ brackets is ``properly nested''. We use three types of opening brackets, $\{ \mathtt{(}, \mathtt{\{}, \mathtt{[}\}$ and corresponding closing brackets $\{ \mathtt{)}, \mathtt{\}}, \mathtt{]}\}$. While the meaning of ``properly nested'' can be grasped intuitively, it can be made more rigorous. We call a sequence ``properly nested'' if it can be generated by the context-free grammar with the following production rule,
\begin{equation}
    \boldsymbol{x}
    \to
    \varepsilon
    \mid
    \boldsymbol{x}\boldsymbol{x}
    \mid
    \texttt{(}
        \boldsymbol{x}
    \texttt{)}
    \mid
    \texttt{\{}
        \boldsymbol{x}
    \texttt{\}}
    \mid
    \texttt{[}
        \boldsymbol{x}
    \texttt{]}
    ,
\end{equation}
where $\boldsymbol{x}$ is a non-terminal symbol, $\varepsilon$ denotes the empty sequence, $\mid$ separates different production rules, and juxtaposition of two symbols denotes concatenation. Equivalently, we could say that the string is an element of the Dyck language with three bracket types~\cite{autebert1997_contextfree, strobl2024_whatformallanguages}.

Algorithm~\ref{alg:brackets-generator} provides pseudocode for the procedure  that we use to create random sequences of brackets that are, in expectation, properly nested in $50\%$ of cases. We use $\mathrm{U}\mathcal{S}$ to denote the uniform distribution on the set $\mathcal{S}$, and make use of the function $\mathtt{get\_closing}$ that associates with an opening bracket the closing bracket of the same type. Throughout, we use the convention that the method $\mathrm{pop}$ returns the empty sequence $\varepsilon$ if called on an empty stack. We set the probability $p_\mathrm{open}=0.6$, slightly favoring deeply nested sequences over ones in which opening and closing brackets alternate, and we set the probability $p_\mathrm{false}=0.5$, thereby obtaining an approximately balanced dataset of sequences of properly and improperly nested brackets. While creating improperly nested sequences by deleting a bracket means that these instances contain only $n - 1$ brackets, thus creating a potential shortcut, we provide evidence in Appendix~\ref{app:brackets} that models are not exploiting this shortcut. We record during sequence generation whether the sequence is or is not properly nested as the binary correct answer $a$.

\begin{algorithm}[tb]
\caption{Generate \Brackets{} sequence}\label{alg:brackets-generator}
\KwData{$n$ (even)}
\KwResult{$\mathbf{x}$, $a\in \{0,1\}$}
$n_\mathrm{open} \leftarrow 0$\;
$n_\mathrm{closed} \leftarrow 0$\;
$\mathbf{x} \leftarrow \mathtt{init}\_\mathtt{list}()$\;
$\mathbf{s} \leftarrow \mathtt{init}\_\mathtt{stack}()$\;
\While{$n_\mathrm{closed} < \frac{n}{2}$}{
    $r \sim \mathrm{U}[0,1]$\;
    \eIf{$n_\mathrm{open}<\frac{n}{2} \wedge \left(r < p_\mathrm{open} \vee  \mathbf{s} =\emptyset \right)$}{
        $o \sim \mathrm{U}\{\texttt{(}, \texttt{\{}, \texttt{[}\}$\;
        $\mathbf{x}\mathrm{.append}(o)$\;
        $c \leftarrow \mathtt{get\_closing}(o)$\;
        $\mathbf{s}\mathrm{.push}(c)$\;
        $n_\mathrm{open} \leftarrow n_\mathrm{open} + 1$\;
    }{
    $c \leftarrow \mathbf{s}\mathrm{.pop}()$\;
    $\mathbf{x}\mathrm{.append}(c)$\;
    $n_\mathrm{closed} \leftarrow n_\mathrm{closed} + 1$\;
  }
}
$r \sim \mathrm{U}[0,1]$\;
$a \leftarrow 1$\;
\If{$r < p_{\mathrm{false}}$}
{
    $i \sim \mathrm{U}\{0,\dots,n-1\}$\;
    $\mathbf{x} \leftarrow \mathbf{x}\mathrm{.delete}(i)$\;
    $a \leftarrow 0$\;
}
\end{algorithm}

A similar stack-based algorithm exists for deciding whether a given sequence $\mathbf{x}$ of brackets is properly nested. We provide pseudocode in Algorithm~\ref{alg:brackets-check}. This algorithm has time complexity $\mathcal{O}(n)$ and uses, in the worst case, $\mathcal{O}(n)$ space for the stack. 

Prior work~\cite{schnabel2025_losttransmission, tomlinson2026_ReasoningReasoningBAPO} does not establish \BAPO{} complexity for \Brackets{}. We conjecture that \Brackets{} is \BAPO{}-hard, but rigorously establishing this is beyond the scope of this paper.

\begin{algorithm}[tb]
\caption{Check \Brackets{} sequence}\label{alg:brackets-check}
\KwData{$\mathbf{x}$}
\KwResult{$a \in \{0,1\}$}
$n \leftarrow \mathtt{len}(\mathbf{x})$\;
$\mathbf{s} \leftarrow \mathtt{init}\_\mathtt{stack}()$\;
$i \leftarrow 0$\;
$z \leftarrow 1$\;
\While{$i < n$}{
    \eIf{$x_i \in \left\{\texttt{(}, \texttt{\{}, \texttt{[}\right\}$}{
        $c \leftarrow \mathtt{get\_closing}(x_i)$\;
        $\mathbf{s}\mathrm{.push}(c)$\;
    }{
        $c \leftarrow \mathbf{s}\mathrm{.pop}()$\;
        \If{$c \neq x_i \vee c = \varepsilon $}{
            $z \leftarrow 0$\;
            \textbf{break}\;
        }
    }
    $i \leftarrow i+1$\; 
}
\eIf{$\mathbf{s} = \emptyset ~\wedge~ z\neq 0$}{
    $a \leftarrow 1$\;
}{
    $a \leftarrow 0$\;
}
\end{algorithm}

\subsubsection{Parity}

A size-$n$ instance of the \Parity{} task consists of a binary string $\mathbf{x}$ of length $n$. We generate random instances of this task by drawing each element of the string uniformly at random from $\{0, 1\}$. The correct answer $a$ can be computed using addition modulo $2$,
\begin{equation}
    a
    =
    \left(
        \sum_{x \in \mathbf{x}}
        x
    \right)
    \mod 2
    .
\end{equation}
An efficient algorithm for this task maintains a parity bit and sequentially updates the parity value as the sequence $\mathbf{x}$ is traversed. Such an algorithm uses $\mathcal{O}(1)$ space for the parity bit and has time complexity $\Theta(n)$, assuming that the parity bit can be updated in $\mathcal{O}(1)$ time. As each bit needs to be touched exactly once, no faster sequential algorithm exists.

In contrast to the \Addition{} task, the \Parity{} task is \BAPO{}-easy, as shown in Example A.1 of Tomlinson et al.~\cite{tomlinson2026_ReasoningReasoningBAPO}.

\subsubsection{Index}

A size-$n$ instance of the \Index{} task consists of a length-$n$ sequence $\mathbf{x}$ of integers and an additional integer $i$ satisfying $0\leq i < n$. We generate random instances of this task by sampling the elements of $\mathbf{x}$ uniformly at random with replacement from $\{0, 1, \dots, 100\}$ and sampling $i$ uniformly at random from $\{0, 1, \dots, n-1\}$. We determine the correct answer $a=x_i$ via lookup.

As the elements of the sequence are of bounded size, this task can be solved using $\mathcal{O}(1)$ space. The time complexity depends on the type of data structure that stores the sequence and ranges (for reasonable choices) from $\mathcal{O}(1)$ look-up for an array to $\mathcal{O}(n)$ traversal for a linked list.

Example A.2 of Tomlinson et al.~\cite{tomlinson2026_ReasoningReasoningBAPO} establishes that \Index{} is \BAPO{}-easy.

\subsection{Extracting Answers}\label{sec:methods:answer-extraction}

Here, we describe how we extract the answer $\hat{a}_{N,n, \tau,r}$ that a model of size $N$ provides for problem instance $r$ of size $n$ at temperature $\tau$ from the output of the model. We use the term ``output'' to refer to all tokens generated after the \texttt{<think>} token including the CoT.

To extract the model answer, we first verify that the CoT was completed, i.e., that the model output contains the token \texttt{</think>}. If the CoT was not completed, we count the instance as incorrectly answered. In the notation of Eq.~\eqref{eq:correct_set}, this can be represented by setting $\hat{a}_{N,n,\tau, r} = \mathtt{NaN}$, where $\mathtt{NaN}$ is a dummy value that does not equal any number. We only consider the part of the output following the CoT, i.e., after the \texttt{</think>} token when extracting the answer.

As the example prompts in Tab.~\ref{tab:tasks} show, we instruct models to enclose their final answers in \texttt{\textbackslash boxed\{\}}. When inspecting the model output, we observe that models obey this instruction in many cases but occasionally use \texttt{\textbackslash{}boxed\{\textbackslash{}text\{\}\}} to enclose answers. We thus use regular expressions to extract possible answers enclosed by either  \texttt{\textbackslash boxed\{\}} or \texttt{\textbackslash{}boxed\{\textbackslash{}text\{\}\}}. The exact regular expression depends on the task. We permit arbitrary whitespace between the enclosing brackets and the answer, and consider the first matching substring as the model's answer.

For the \Addition{} and \Index{} tasks we match any integer and account for possible digit-grouping commas, i.e., both \texttt{1000} and \texttt{1,000} will be recognized as valid answers. For the \Brackets{} task, we match \texttt{True} or \texttt{False} using case-insensitive matching. Finally, for the \Parity{} task, we match \texttt{even} or \texttt{odd}, again, using case-insensitive matching.

\subsection{Statistical Models}\label{sec:methods:statistical-models}

\subsubsection{Variable Asymptote Model}

Here, we provide the full formulation of the variable asymptote model (VAM), including all priors and hyperpriors.

We use the binomial likelihood for the observed number of instances $y_{N,n,\tau}$ of size $n$ that a model of size $N$ solves correctly at temperature $\tau$ out of $R$ total problem instances,
\begin{equation}
    y_{N,n,\tau}
    \sim
    \mathrm{Binom}
    \left(
        R, p_{N,\tau}(n)
    \right).
\end{equation}
The function $p_{N,\tau}(n)$ is given in Eq.~\eqref{eq:exponential-decay-p} and requires us to specify the parameters $p_\tau^{(0)}$, $p_{N,\tau}^{(\infty)}$, and $\nu_{N,\tau}$.

To ensure $p^{(0)}_{\tau}$ and $p_{N,\tau}^{(\infty)}$ are confined to the interval $[0,1]$, we parameterize 
\begin{equation}\label{eq:prior-exponential-decay-p}
\begin{split}
    \kappa_{\tau}^{(0)}
    &\sim
    \mathcal{N}(1.5, 1.0) \\
    \kappa_{N,\tau}^{(\Delta)}
    &\sim
    \mathcal{N}(1.5, 1.0) \\
    p_{\tau}^{(0)}
    &=
    \sigma\left(\kappa_{\tau}^{(0)}\right)\\
    p_{N,\tau}^{(\infty)}
    &=
    p^{(0)}_{\tau} \sigma\left(\kappa_{N,\tau}^{(\Delta)} \right)
    ,
\end{split}
\end{equation}
where $\sigma$ is the sigmoid function, $\sigma(x)=\frac{1}{1+e^{-x}}$, and $\mathcal{N}(\tilde{\mu},\tilde{\sigma})$ denotes the normal distribution with mean $\tilde{\mu}$ and standard deviation $\tilde{\sigma}$.

The model capability parameter $\nu_{N,\tau}$ is defined through the latent scaling relationship in Eq.~\eqref{eq:capability-scaling} that depends on the parameters $C_{\tau}^{(\nu)}$, $\beta_{\tau}^{(\nu)}$, $\eta_{\tau}^{(\nu)}$, and the noise term $\xi_{N,\tau}^{(\nu)}$. We use the following priors
\begin{equation}
\begin{split}
    \log C_{\tau}^{(\nu)}
    &\sim
    \mathcal{N}(\log 50, 1) \\
    \beta_{\tau}^{(\nu)}
    &\sim
    \mathcal{N}(0, 1) \\ 
    \log \eta_{\tau}^{(\nu)}
    &\sim
    \mathcal{N}(\log 0.3, 1)\\
    \xi_{N,\tau}^{(\nu)}
    &\sim
    \mathcal{N}(0,1)
\end{split}
\end{equation}
where sampling on a logarithmic scale is employed to ensure that the prefactor $C_{\tau}^{(\nu)}$ and noise magnitude $\eta^{(\nu)}_{\tau}$ are positive.

\subsubsection{Fixed Asymptote Model}

The fixed asymptote model (FAM) uses the same priors as the VAM, with one modification that prevents $p_{N,\tau}^{(\infty)}$ from varying with $N$.

Instead of sampling $p_{N,\tau}^{(\infty)}$ independently for each value of $N$ at a fixed temperature $\tau$ as in Eq.~\eqref{eq:prior-exponential-decay-p}, we sample one parameter $p_{\tau}^{(\infty)}$ per value of the temperature $\tau$ with prior
\begin{equation}
\begin{split}
    \kappa_{\tau}^{(\Delta)}
    &\sim
    \mathcal{N}(1.5, 1.0)\\
    p_{\tau}^{(\infty)}
    &=
    p_{\tau}^{(0)}\sigma\left(\kappa_{\tau}^{(\Delta)}\right).
\end{split}
\end{equation}
The analogue of the exponential decay Eq.~\eqref{eq:exponential-decay-p} is 
\begin{equation}
    p_{N,\tau}(n)
    =
    p_{\tau}^{(\infty)}
    +
    \left(
        p_{\tau}^{(0)}
        -
        p_{\tau}^{(\infty)}
    \right)
    \exp
    \left(
        -\frac{n}{\nu_{N,\tau}}
    \right)
    .
\end{equation}

The parametric assumptions on $\nu_{N,\tau}$ are identical in both models.

\subsubsection{Prefactor Exponent Model}

The prefactor exponent model (PEM) describes the average output length $\bar{\ell}_{N,n,\tau}$ (Eq.~\eqref{eq:avg_len_correct}) of a correct answer using a Gaussian likelihood for the logarithm,
\begin{equation}
    \log \bar{\ell}_{N,n,\tau}
    \sim
    \mathcal{N}
    \left(
        \log \lambda_{N,n,\tau},
        \omega_{N,n,\tau}
    \right),
\end{equation}
where $\lambda_{N,n,\tau}$ and $\omega_{N,n,\tau}$ are given by Eq.~\eqref{eq:avg_len_scaling_mean} and Eq.~\eqref{eq:avg_len_scaling_var}, respectively. The latter contains the parameter $\zeta_{\tau}$ for which we use a half-normal prior
\begin{equation}
    \zeta_{\tau}
    \sim 
    \mathrm{HalfNorm}(0.5)
    .
\end{equation}

The former, Eq.~\eqref{eq:avg_len_scaling_mean}, depends on the parameters $A_{N,\tau}$ and $\alpha_{N,\tau}$ that are specified by the latent scaling relationships in Eq.~\eqref{eq:len_prefactor_scaling} and Eq.~\eqref{eq:len_exponent_scaling}, respectively. We need to specify priors for all parameters in these equations, i.e., for the prefactors $C^{(A)}_{\tau}$, $C^{(\alpha)}_{\tau}$, 
\begin{equation}
\begin{split}
    \log C^{(A)}_{\tau}
    &\sim 
    \mathcal{N}(\log 10^4, 0.75)\\
    \log C^{(\alpha)}_{\tau}
    &\sim
    \mathcal{N}(\log 0.75, 0.35)
\end{split}
\end{equation}
the scaling exponents $\beta^{(A)}_{\tau}$, $\beta^{(\alpha)}_{\tau}$,
\begin{equation}\label{eq:len_scaling_priors_beta}
\begin{split}
    \beta_{\tau}^{(A)}
    &\sim 
    \mathcal{N}(0, 0.5)\\
    \beta_{\tau}^{(\alpha)}
    &\sim
    \mathcal{N}(0, 0.5)\\
\end{split}   
\end{equation}
and the noise terms $\eta_{\tau}^{(A)} \xi^{(A)}_{N,\tau}$, $\eta^{(\alpha)}_{\tau} \xi^{(\alpha)}_{N,\tau}$ have priors
\begin{equation}\label{eq:len_scaling_priors}
\begin{split}
    \eta^{(A)}_{\tau}
    &\sim
    \mathrm{HalfNorm}(0.25)\\
    \eta^{(\alpha)}_{\tau}
    &\sim
    \mathrm{HalfNorm}(0.25)\\
    \xi_{N,\tau}^{(A)}
    &\sim
    \mathcal{N}(0,1)\\
    \xi_{N,\tau}^{(\alpha)}
    &\sim
    \mathcal{N}(0, 1)
    .
\end{split}
\end{equation}

\subsubsection{Prefactor Model}

The prefactor model (PM) uses the same priors as the PEM but does not include latent scaling of the exponent, $\alpha_{N,\tau}$. Thus, Eq.~\eqref{eq:len_exponent_scaling} simplifies to
\begin{equation}\label{eq:len_exponent_no_scaling}
    \log \alpha_{N,\tau}
    =
    \log C_{\tau}^{(\alpha)}
    +
    \eta_{\tau}^{(\alpha)} \xi^{(\alpha)}_{N,\tau},
\end{equation}
and all variation across different values of $N$ is due to the random effect term. As $\beta^{(\alpha)}_{\tau}$ does not appear in Eq.~\eqref{eq:len_exponent_no_scaling}, we can omit the relevant prior in Eq.~\eqref{eq:len_scaling_priors_beta}. 

\subsection{HMC Hyperparameters and Diagnostics}\label{sec:methods:MCMC}

To infer the parameters of our statistical models, we perform Hamiltonian Monte Carlo (HMC) sampling using the No-U-Turn Sampler (NUTS) implemented in the \texttt{NumPyro} library~\cite{bingham2019_pyrodeepuniversal, phan2019_composableeffectsflexible}. We run six independent chains, each with $1{,}000$ warm-up iterations and $5{,}000$ post-warm-up samples. We use a target acceptance probability of $0.99$ and a maximum tree depth of $20$, and otherwise use the default configuration.

When fitting the \FAM{}, \VAM{}, \PM{}, and \PEM{} on the full dataset results we encountered no divergences and found $\hat{R} <  1.01$ throughout.

To assess out-of-sample predictive performance, we estimate the leave-one-out expected log predictive density ($\elpd{}$). At each temperature $\tau$, we leave out one combination of model size $N$ and instance size $n$, refit the statistical model to the remaining data, and evaluate its predictive density on the held-out configuration.

When comparing the \FAM{} and \VAM{}, we did not encounter divergences and found $\hat{R} < 1.01$ for all fits. 
We also did not encounter any divergences and found $\hat{R}<1.01$ for all fits of the \PM{} and \PEM{}.

\section{Code Availability}

We make code for generating reasoning traces, as well as for our statistical analysis available at~\url{https://github.com/zohairshafi/capability-efficiency-scaling}.

\section{Acknowledgments}

M.L.~and T.E.R.~are supported by the Inaugural Joseph E.~Aoun Endowment. We thank Modal Labs, Inc. for providing compute resources for this project.

\section{AI Use}

We used the following AI assistants for proofreading the manuscript: OpenAI's GPT-5.5-Sol and GPT-5.6-Sol and GPT-6-Astra and Anthropic's Fable 5 and Fable 5.1. We used OpenAI's GPT-5.6-Sol to aide in mathematical derivations, which we verified. We used OpenAI's GPT-5.6-Sol and DeepSeek AI's DeepSeek-v4-Pro with Anthropic's Claude Code harness in the software development process. We validated the all code to ensure correctness.

%%%%%%%%%%%%%%%%%%%%%%%%%%%%%%%%%%%%%%%%%%%%%
%             REFERENCES                    %
%%%%%%%%%%%%%%%%%%%%%%%%%%%%%%%%%%%%%%%%%%%%%%

\renewcommand{\bibsection}{\section*{References}}
\bibliographystyle{mybst}
\bibliography{bib}

\clearpage
\appendix
\onecolumngrid
\setcounter{figure}{0}
\setcounter{table}{0}
\renewcommand{\thefigure}{A\arabic{figure}}
\renewcommand{\thetable}{A\arabic{table}}

% --- SI title block ---
\onecolumngrid
\phantomsection

\begin{center}
{\large\bfseries Appendices for}\\[0.35em]
    {\large\bfseries Scaling of Capability \& Efficiency at Inference Time in Large Reasoning Models}\\[0.75em]

{\normalsize
Moritz Laber,
Zohair Shafi,
Germans Savcisens,
Brennan Klein,
Matteo Chinazzi,
Samuel V. Scarpino,
Albert-László Barabási,
Alessandro Vespignani,
Tina Eliassi-Rad
}\\[0.35em]

{\small \texttt{\url{laber.m@northeastern.edu}}}
\end{center}

\section{Likelihood Variance in PM and PEM}\label{app:delta-method}

For ease of notation, we drop the subscripts $N,n,\tau$ in this section. 
Let $\{L_r\}_{r=1}^{y}$ be a collection of random variables, each with mean $\mu_L$ and variance $\sigma_L^2$. Then their arithmetic mean
\begin{equation}
    \bar{L}
    =
    \frac{1}{y}
    \sum_{r=1}^{y} L_r
\end{equation}
is itself a random variable with expected value
\begin{equation}
    \mathbb{E}
    \left[
        \bar{L}
    \right]
    =
    \mathbb{E}
    \left[
        \frac{1}{y}
        \sum_{r=1}^y
        L_r
    \right]
    =
    \frac{1}{y}
    \sum_{r=1}^{y}
    \mathbb{E}[L_r]
    =
    \frac{1}{y}
    \sum_{r=1}^y
    \mu_L
    =
    \mu_L
\end{equation}
and variance
\begin{equation}
    \mathrm{Var}
    \left[
        \bar{L}
    \right]
    =
    \mathrm{Var}
    \left[
        \frac{1}{y}
        \sum_{r=1}^{y}
        L_r
    \right]
    =
    \frac{1}{y^2}
    \mathrm{Var}
    \left[
        \sum_{r=1}^{y}
        L_r
    \right]
    =
    \frac{1}{y^2}
    \sum_{r=1}^{y}
    \mathrm{Var}
    \left[
        L_r
    \right]
    =
    \frac{\sigma_L^2}{y}
    .
\end{equation}

If we observe $f(\bar{L})$ for a sufficiently well-behaved function $f$, close to the expected value of $\bar{L}$, i.e., $\bar{L}=\mu_L + \delta$ with $\mathbb{E}[\delta]=0$ and $\mathrm{Var}[\delta]=\frac{\sigma_L^2}{y}$, we can use a first-order Taylor expansion to derive the mean
\begin{equation}
    \mathbb{E}
    \left[
        f(\bar{L})
    \right]
    =
    \mathbb{E}
    \left[
        f(\mu_L +\delta)
    \right]
    \approx
    \mathbb{E}
    \left[
        f(\mu_L)
    \right]
    +
    f'(\mu_L)
    \mathbb{E}
    \left[
        \delta
    \right]
    =
    f(\mu_L)
\end{equation}
and variance
\begin{equation}
    \mathrm{Var}
    \left[
        f(\bar{L})
    \right]
    =
    \mathrm{Var}
    \left[
        f(\mu_L + \delta)
    \right]
    \approx
    \mathrm{Var}
    \left[
        f(\mu_L) + f'(\mu_L)\delta
    \right]
    =
    f'(\mu_L)^2
    \mathrm{Var}
    \left[
        \delta
    \right]
    =
    f'(\mu_L)^2
    \frac{\sigma_L^2}{y}
    ,
\end{equation}
where $f'$ denotes the first derivative of $f$. 

In the \PM{} and \PEM{}, the relevant function $f$ is the natural logarithm and hence
\begin{equation}
    \mathbb{E}
    \left[
        \log \bar{L}
    \right]
    \approx
    \log \mu_L
\end{equation}
and
\begin{equation}
    \mathrm{Var}
    \left[
        \log \bar{L}
    \right]
    \approx
    \frac{1}{\mu_L^2}
    \frac{\sigma^2}{y},
\end{equation}

as $\frac{\mathrm{d}}{\mathrm{d}x}\log x = \frac{1}{x}$.

\section{Additional Results}\label{app:additional-results}

This Appendix presents additional results on the scaling of capability and efficiency in models from the \ROne{} model family.

\subsection{Capability}\label{app:additional-results:capability}

\begin{table}[tb]
\begin{tabular}{|l|c|c|c|c|c|c|}
\hline
task & $\tau$ & stat. model & $\postE{\beta_{\tau}^{(\nu)}}$ & $\CrI{95}[\beta_{\tau}^{(\nu)}]$ & $\postE{\eta_{\tau}^{(\nu)}}$ & $\CrI{95}[\eta_{\tau}^{(\nu)}]$  \\ \hhline{|=|=|=|=|=|=|=|}
\Addition{}  & 0.4 & \FAM{} & 0.68 & [0.51, 0.85] & 0.22 & [0.10, 0.47] \\
\Addition{}  & 0.6 & \FAM{} & 0.69 & [0.53, 0.85] & 0.20 & [0.08, 0.44] \\
\Addition{}  & 0.8 & \FAM{} & 0.70 & [0.52, 0.87] & 0.23 & [0.10, 0.47] \\ \hline
\Brackets{}  & 0.4 & \FAM{} & 0.65 & [0.32, 0.97] & 0.45 & [0.26, 0.80] \\
\Brackets{}  & 0.6 & \FAM{} & 0.69 & [0.31, 1.06] & 0.52 & [0.30, 0.91] \\
\Brackets{}  & 0.8 & \FAM{} & 0.57 & [0.27, 0.86] & 0.36 & [0.17, 0.70] \\ \hline
\Index{}     & 0.4 & \FAM{} & 0.58 & [0.24, 0.91] & 0.48 & [0.27, 0.84] \\
\Index{}     & 0.6 & \FAM{} & 0.52 & [0.19, 0.86] & 0.47 & [0.27, 0.82] \\
\Index{}     & 0.8 & \FAM{} & 0.51 & [0.18, 0.83] & 0.45 & [0.26, 0.80] \\ \hline
\Parity{}    & 0.4 & \FAM{} & 0.75 & [0.38, 1.11] & 0.51 & [0.29, 0.88] \\
\Parity{}    & 0.6 & \FAM{} & 0.69 & [0.40, 0.98] & 0.40 & [0.22, 0.72] \\
\Parity{}    & 0.8 & \FAM{} & 0.73 & [0.41, 1.04] & 0.43 & [0.24, 0.78] \\ \hline
\Addition{}  & 0.4 & \VAM{} & 0.60 & [0.41, 0.80] & 0.25 & [0.11, 0.51] \\
\Addition{}  & 0.6 & \VAM{} & 0.58 & [0.39, 0.77] & 0.24 & [0.11, 0.49] \\
\Addition{}  & 0.8 & \VAM{} & 0.61 & [0.40, 0.82] & 0.27 & [0.12, 0.53] \\ \hline
\Brackets{}  & 0.4 & \VAM{} & 0.26 & [0.06, 0.46] & 0.25 & [0.11, 0.51] \\
\Brackets{}  & 0.6 & \VAM{} & 0.07 & [-0.17, 0.30] & 0.29 & [0.12, 0.58] \\
\Brackets{}  & 0.8 & \VAM{} & 0.28 & [0.00, 0.55] & 0.33 & [0.13, 0.66] \\ \hline
\Index{}     & 0.4 & \VAM{} & 0.49 & [0.14, 0.84] & 0.48 & [0.27, 0.86] \\
\Index{}     & 0.6 & \VAM{} & 0.42 & [0.06, 0.78] & 0.51 & [0.27, 0.90] \\
\Index{}     & 0.8 & \VAM{} & 0.42 & [0.11, 0.72] & 0.42 & [0.21, 0.77] \\ \hline
\Parity{}    & 0.4 & \VAM{} & 0.34 & [0.01, 0.67] & 0.45 & [0.25, 0.80] \\
\Parity{}    & 0.6 & \VAM{} & 0.25 & [-0.07, 0.57] & 0.44 & [0.24, 0.79] \\
\Parity{}    & 0.8 & \VAM{} & 0.31 & [-0.04, 0.67] & 0.47 & [0.25, 0.84] \\ \hline
\end{tabular}
\caption{\textbf{Capability scaling.} The posterior means of the scaling exponents $\beta_{\tau}^{(\nu)}$ support a sublinear latent scaling for all tasks and temperatures, irrespective of the statistical model used. We also show $95\%$ credible intervals. The posterior mean and $95\%$ credible intervals of $\eta_{\tau}^{(\nu)}$ indicate that the deviations from the latent scaling law are relatively small throughout.}
\label{tab:capability-scaling}
\end{table}

\begin{table}[tb]
    \centering
\begin{tabular}{|l|c|c|c|c|c|c|c|}
\hline
\multicolumn{3}{|c}{~} & \multicolumn{5}{|c|}{$\postE{\nu_{N,\tau}}$} \\ \hline
 task & $\tau$ & stat. model & $1.5\B$ & $7\B$ & $14\B$ & $32\B$ & $70\B$\\ \hhline{|=|=|=|=|=|=|=|=|}
\Addition{} & 0.4 & \FAM{} & 22 & 68  & 124 & 208 & 287 \\
\Addition{} & 0.6 & \FAM{} & 26 & 67  & 131 & 212 & 363 \\
\Addition{} & 0.8 & \FAM{} & 26 & 64  & 140 & 222 & 362 \\ \hline
\Brackets{} & 0.4 & \FAM{} & 24 & 54  & 244 & 161 & 299 \\
\Brackets{} & 0.6 & \FAM{} & 32 & 62  & 390 & 302 & 362 \\ 
\Brackets{} & 0.8 & \FAM{} & 8  & 25  & 44 & 85   & 66 \\ \hline
\Index{}    & 0.4 & \FAM{} & 58 & 186 & 540 & 253 & 685 \\
\Index{}    & 0.6 & \FAM{} & 75 & 202 & 615 & 287 & 679 \\
\Index{}    & 0.8 & \FAM{} & 81 & 201 & 642 & 329 & 646 \\ \hline
\Parity{}   & 0.4 & \FAM{} & 23 & 80  & 285 & 472 & 294 \\ 
\Parity{}   & 0.6 & \FAM{} & 30 & 97  & 292 & 369 & 366 \\
\Parity{}   & 0.8 & \FAM{} & 25 & 103 & 296 & 369 & 353 \\ \hline
\Addition{} & 0.4 & \VAM{} & 22 & 67  & 120 & 166 & 226 \\
\Addition{} & 0.6 & \VAM{} & 26 & 66  & 127 & 158 & 250 \\
\Addition{} & 0.8 & \VAM{} & 26 & 63  & 136 & 148 & 293 \\ \hline
\Brackets{} & 0.4 & \VAM{} & 22 & 44  & 45  & 60  & 61 \\
\Brackets{} & 0.6 & \VAM{} & 27 & 42  & 39  & 45  & 34 \\
\Brackets{} & 0.8 & \VAM{} & 12 & 26  & 27  & 45  & 32 \\ \hline
\Index{}    & 0.4 & \VAM{} & 57 & 175 & 404 & 174 & 537 \\
\Index{}    & 0.6 & \VAM{} & 74 & 194 & 421 & 157 & 538 \\
\Index{}    & 0.8 & \VAM{} & 81 & 184 & 392 & 202 & 528 \\ \hline
\Parity{}   & 0.4 & \VAM{} & 18 & 73  & 50  & 41  & 110 \\
\Parity{}   & 0.6 & \VAM{} & 25 & 81  & 57  & 42  & 103 \\
\Parity{}   & 0.8 & \VAM{} & 16 & 74  & 61  & 41  & 80 \\
\hline
\end{tabular}
\caption{\textbf{Capability parameters.} The model capability, quantified by the posterior mean of the scale $\nu_{N,\tau}$ of the exponential decay of the expected number of correctly solved instances as a function of the instance size, generally increases as a function of the model size $N$. However, model-specific posterior means can be non-monotonic in some conditions. The parameter $\nu_{N,\tau}$ can be thought of as the characteristic size scale of an instance that is tractable for models of size $N$ at temperature $\tau$.}
\label{tab:capability-coefficients}
\end{table}

Table~\ref{tab:capability-scaling} summarizes the results of our analysis of capability scaling for all reasoning tasks, temperatures, and statistical models. The inferred exponential decay scales are documented in Tab.~\ref{tab:capability-coefficients}.

For the \Addition{} task, we find sublinear scaling $\postE{\beta_{\tau}^{(\nu)}} < 1$ for both statistical models and at all temperatures. The expected size of deviations from the latent scaling law, $\postE{\eta_{\tau}^{(\nu)}}$, is small throughout. We find that the typical size of a solvable instance increases from ca. $20$ at $N=1.5\B$ parameters to several hundred at $N=70\B$ parameters. Model comparison favors the \FAM{} over the \VAM{} at all temperatures (Tab.~\ref{tab:model-comparison}).

Turning to the \Brackets{} task, we find that the \FAM{} infers higher scaling exponents than the \VAM{}. The latter is favored in model comparison for all temperatures (Tab.~\ref{tab:model-comparison}), and shows lower deviations from the latent scaling law. Both statistical models agree that capability scaling is sublinear. 

On the \Index{} task, the \FAM{} is favored over the \VAM{} in model comparison (Tab.~\ref{tab:model-comparison}) at all temperatures. Both statistical models show sublinear scaling with similar scaling exponents. The deviations from the latent scaling law are moderate and of similar size for both statistical models. 

Finally, for the \Parity{} task, scaling is inferred to be sublinear but the \FAM{} again infers higher scaling exponents than the \VAM{}. However, the latter is preferred in model comparison (Tab.~\ref{tab:model-comparison}). Deviations from the latent scaling law are comparable in both cases. 

In summary, these results support sublinear scaling of model capability with model size.

\subsection{Efficiency}\label{app:additional-results:efficiency}

Table~\ref{tab:efficiency-scaling} provides an overview of our analysis of efficiency scaling for all reasoning tasks, temperatures, and statistical models.

Our analysis of the \Addition{} task using the \PM{} shows that the posterior mean of the scaling exponent of the prefactor $A_{N,\tau}$ is close to zero, $|\postE{\beta_{\tau}^{(A)}}|\leq 0.1$. The \PEM{}, which is preferred in model comparison, infers weakly negative expected scaling exponents for the prefactor $A_{N,\tau}$ and the exponent $\alpha_{N,\tau}$ of the power law linking instance size and output length. However, the $95\%$ credible interval covers both negative and positive values.

For the \Brackets{} task, we find slightly positive scaling exponents for the prefactor $A_{N,\tau}$ at temperatures $\tau \in \{0.4, 0.6\}$, and an exponent close to zero at $\tau = 0.8$ using the \PM{}. Similar values are obtained using the \PEM{}. The former statistical model is preferred in model comparison. Using the \PEM{}, we find scaling exponents for the exponent $\alpha_{N,\tau}$ close to zero, $|\postE{\beta_{\tau}^{(\alpha)}}|\leq 0.06$. 

When it comes to the \Index{} task, both the \PM{} and the \PEM{} indicate expected scaling exponents of the prefactor $A_{N,\tau}$ of small magnitude, $|\postE{\beta_{\tau}^{(A)}}| \leq 0.05$. The \PEM{} also suggests a very weak dependence of the exponent $\alpha_{N,\tau}$ on the model size $N$, $|\postE{\beta_{\tau}^{(\alpha)}}|\leq 0.07$. The $95\%$ credible intervals for these parameters cover both positive and negative values. Model comparison favors the \PEM{} at temperature $\tau = 0.8$, and the \PM{} at lower temperatures $\tau \in \{0.4, 0.6\}$.

Finally, for the \Parity{} task both the \PM{} and the \PEM{} suggest that the expected scaling exponent of the prefactor $A_{N,\tau}$ is slightly negative, $-0.17 \leq \postE{\beta_{\tau}^{(A)}} \leq -0.12$. However, the $95\%$ credible interval contains both positive and negative values. The \PEM{} does not provide strong evidence for scaling of the exponent $\alpha_{N,\tau}$ with model size $N$, as $|\postE{\beta_{\tau}^{(\alpha)}}| \leq 0.06$.

Turning to the actually inferred values of the prefactor $A_{N,\tau}$ and exponent $\alpha_{N,\tau}$ (Tab.~\ref{tab:efficiency-coefficients}), we find that the posterior mean $\postE{\alpha_{N,\tau}}$ of the exponent ranges from $0.41$ to $0.89$. This suggests a sublinear scaling of output length with the instance size. This result is somewhat surprising given the time and space complexities documented in Tab.~\ref{tab:tasks}. However, it is worth noting that reasoning tokens do not directly translate into computational steps. For example, in the \Addition{} task a model could compute the sum of several numbers without explicitly stating this computation, thereby achieving sublinear output length scaling. For the \Index{} task, on the other hand, $\Theta(1)$ time complexity requires special data structures that the model might not be able to exploit internally. The inferred prefactors $A_{N,\tau}$ are on average (across model sizes and temperatures) largest for the \Addition{} task, and smallest for the \Index{} task. This pattern holds irrespective of the statistical model.

Together, these results support the claim that efficiency scaling, formalized as a dependence of the prefactor $A_{N,\tau}$ on the model size $N$, is weak or absent in the \ROne{} model family applied to our four reasoning tasks. 

\begin{table}[tb]
\centering
\begin{tabular}{|l|c|c|c|c|c|c|c|c|c|c|}
\hline
 task & $\tau$ & stat. model & $\postE{\beta_{\tau}^{(A)}}$ & $\CrI{95}[\beta_{\tau}^{(A)}]$ & $\postE{\eta_{\tau}^{(A)}}$ & $\CrI{95}[\eta_{\tau}^{(A)}]$ & $\postE{\beta_{\tau}^{(\alpha)}}$ & $\CrI{95}[\beta_{\tau}^{(\alpha)}]$ & $\postE{\eta_{\tau}^{(\alpha)}}$ & $\CrI{95}[\eta_{\tau}^{(\alpha)}]$ \\
\hhline{|=|=|=|=|=|=|=|=|=|=|=|}
\Addition{} & 0.4 & \PM{}  & -0.05 & [-0.44, 0.35] & 0.53  & [0.33, 0.79] & -     & -              & 0.25 & [0.10, 0.48] \\
\Addition{} & 0.6 & \PM{}  & -0.07 & [-0.50, 0.41] & 0.51  & [0.30, 0.78] & -     & -              & 0.26 & [0.10, 0.50] \\
\Addition{} & 0.8 & \PM{}  & -0.08 & [-0.51, 0.41] & 0.55  & [0.33, 0.82] & -     & -              & 0.30 & [0.15, 0.54] \\ \hline
\Brackets{} & 0.4 & \PM{}  &  0.15 & [0.00, 0.29] & 0.13  & [0.01, 0.38] & -     & -              & 0.11 & [0.00, 0.32] \\
\Brackets{} & 0.6 & \PM{}  &  0.12 & [-0.03, 0.25] & 0.14 & [0.01, 0.38] & -     & -              & 0.08 & [0.00, 0.25] \\
\Brackets{} & 0.8 & \PM{}  & -0.02 & [-0.16, 0.12] & 0.15 & [0.01, 0.38] & -     & -              & 0.10 & [0.00, 0.29] \\
\hline
\Index{}    & 0.4 & \PM{}  & -0.05 & [-0.34,  0.25] & 0.43 & [0.25, 0.69] & -     & -              & 0.08 & [0.00, 0.25] \\
\Index{}    & 0.6 & \PM{}  & -0.03 & [-0.32,  0.25] & 0.42 & [0.24, 0.68] & -     & -              & 0.09 & [0.00, 0.28] \\
\Index{}    & 0.8 & \PM{}  &  0.01 & [-0.23,  0.26] & 0.35 & [0.18, 0.60] & -     & -              & 0.20 & [0.04, 0.43] \\
\hline
\Parity{}   & 0.4 & \PM{}  & -0.17 & [-0.53,  0.19] & 0.55 & [0.35, 0.81] & -     & -              & 0.26 & [0.07, 0.50] \\
\Parity{}   & 0.6 & \PM{}  & -0.15 & [-0.50,  0.21] & 0.54 & [0.35, 0.80] & -     & -              & 0.25 & [0.07, 0.49] \\
\Parity{}   & 0.8 & \PM{}  & -0.12 & [-0.45,  0.22] & 0.52 & [0.33, 0.77] & -     & -              & 0.25 & [0.11, 0.48] \\
\hline
\Addition{} & 0.4 & \PEM{} & -0.09 & [-0.49,  0.32] & 0.54 & [0.33, 0.80] & -0.11 & [-0.31,  0.09] & 0.23 & [0.07, 0.48] \\
\Addition{} & 0.6 & \PEM{} & -0.13 & [-0.58,  0.38] & 0.53 & [0.31, 0.81] & -0.12 & [-0.33,  0.12] & 0.25 & [0.08, 0.50] \\
\Addition{} & 0.8 & \PEM{} & -0.14 & [-0.57,  0.35] & 0.56 & [0.35, 0.83] & -0.13 & [-0.35,  0.10] & 0.28 & [0.13, 0.53] \\
\hline
\Brackets{} & 0.4 & \PEM{} &  0.15 & [0.01,   0.29] & 0.12 & [0.00, 0.38] & 0.06  & [-0.07,  0.20] & 0.12 & [0.00, 0.35] \\
\Brackets{} & 0.6 & \PEM{} &  0.12 & [-0.02,  0.25] & 0.14 & [0.01, 0.38] & 0.04  & [-0.06,  0.15] & 0.09 & [0.00, 0.29] \\
\Brackets{} & 0.8 & \PEM{} & -0.02 & [-0.16,  0.12] & 0.15 & [0.01, 0.39] & -0.02 & [-0.15,  0.12] & 0.12 & [0.00, 0.36] \\
\hline
\Index{}    & 0.4 & \PEM{} & -0.05 & [-0.34,  0.25] & 0.43 & [0.26, 0.69] & -0.02 & [-0.14,  0.10] & 0.09 & [0.00, 0.31] \\
\Index{}    & 0.6 & \PEM{} & -0.03 & [-0.32,  0.25] & 0.42 & [0.24, 0.68] & 0.00  & [-0.12,  0.13] & 0.11 & [0.00, 0.34] \\
\Index{}    & 0.8 & \PEM{} &  0.01 & [-0.24,  0.26] & 0.35 & [0.18, 0.60] & 0.07  & [-0.11,  0.25] & 0.21 & [0.03, 0.46] \\
\hline
\Parity{}   & 0.4 & \PEM{} & -0.17 & [-0.53,  0.19] & 0.55 & [0.35, 0.81] & -0.03 & [-0.25,  0.20] & 0.28 & [0.09, 0.55] \\
\Parity{}   & 0.6 & \PEM{} & -0.15 & [-0.50,  0.21] & 0.54 & [0.35, 0.81] & 0.04  & [-0.17,  0.26] & 0.27 & [0.09, 0.53] \\
\Parity{}   & 0.8 & \PEM{} & -0.12 & [-0.45,  0.22] & 0.52 & [0.33, 0.77] & 0.06  & [-0.14,  0.26] & 0.27 & [0.12, 0.51] \\ \hline
\end{tabular}
\caption{\textbf{Efficiency scaling.} The posterior mean of the scaling exponent $\beta_{\tau}^{(A)}$ of the prefactor $A_{N,\tau}$ of the power law linking instance size $n$ and average output length $\bar{\ell}_{N,n,\tau}$ for a correctly solved instance  is close to zero, compatible with the idea that efficiency does not scale with model size. Its $95\%$ credible interval covers both positive and negative values in most cases. These results hold for both statistical models. The \PEM{}, which also allows for scaling of the power law exponent $\alpha_{N,\tau}$, shows low posterior mean values of the associated scaling exponent $\beta_{\tau}^{(\alpha)}$, with a $95\%$ credible interval covering both positive and negative values. The expected values of standard deviations $\eta_{\tau}^{(A)}$ and $\eta_{\tau}^{(\alpha)}$ of the random effects associated with either scaling law tend to be small.}
\label{tab:efficiency-scaling}
\end{table}

\begin{table}[tb]
\centering
\begin{tabular}{|l|c|c||c|c|c|c|c||c|c|c|c|c|}
\hline
\multicolumn{3}{|c||}{~} &   \multicolumn{5}{c||}{$\postE{A_{N,\tau}}$} & \multicolumn{5}{c|}{$\postE{\alpha_{N,\tau}}$} \\ \hline
task & $\tau$ & stat. model & $1.5\B$ & $7\B$ & $14\B$ & $32\B$ & $70\B$ & $1.5\B$ & $7\B$ & $14\B$ & $32\B$ & $70\B$ \\
\hhline{|=|=|=||=|=|=|=|=||=|=|=|=|=|}
 \Addition{} & 0.4 & \PM{}  & 11577 & 2990 & 13029 & 3992 & 8621   & 0.78 & 0.65 & 0.77 & 0.77 & 0.48 \\
 \Addition{} & 0.6 & \PM{}  & 15319 & 3301 & 11628 & 4468 & 9031   & 0.75 & 0.68 & 0.74 & 0.80 & 0.47 \\
 \Addition{} & 0.8 & \PM{}  & 17387 & 3002 & 12952 & 4573 & 9418   & 0.79 & 0.64 & 0.73 & 0.82 & 0.42 \\ \hline
 \Brackets{} & 0.4 & \PM{}  & 3748  & 4677 & 5595  & 5589 & 6905   & 0.57 & 0.63 & 0.65 & 0.63 & 0.62 \\
 \Brackets{} & 0.6 & \PM{}  & 3919  & 4168 & 5346  & 5189 & 6129   & 0.56 & 0.59 & 0.61 & 0.60 & 0.59 \\
 \Brackets{} & 0.8 & \PM{}  & 4607  & 3846 & 3745  & 4017 & 4199   & 0.49 & 0.52 & 0.48 & 0.51 & 0.48 \\ \hline
 \Index{}    & 0.4 & \PM{}  & 4684  & 1631 & 3409  & 1969 & 3951   & 0.55 & 0.54 & 0.56 & 0.54 & 0.53 \\
 \Index{}    & 0.6 & \PM{}  & 4720  & 1820 & 3544  & 2171 & 4372   & 0.53 & 0.55 & 0.57 & 0.57 & 0.52 \\
 \Index{}    & 0.8 & \PM{}  & 4035  & 2120 & 3612  & 2562 & 4360   & 0.43 & 0.58 & 0.61 & 0.62 & 0.50 \\ \hline
 \Parity{}   & 0.4 & \PM{}  & 8902  & 5198 & 4716  & 1376 & 7590   & 0.57 & 0.72 & 0.46 & 0.45 & 0.64 \\
 \Parity{}   & 0.6 & \PM{}  & 8285  & 4890 & 6480  & 1444 & 7653   & 0.46 & 0.70 & 0.50 & 0.45 & 0.63 \\
 \Parity{}   & 0.8 & \PM{}  & 7175  & 5547 & 6968  & 1619 & 7461   & 0.42 & 0.72 & 0.54 & 0.47 & 0.62 \\ \hline
 \Addition{} & 0.4 & \PEM{} & 14435 & 3096 & 13047 & 3987 & 8611   & 0.87 & 0.67 & 0.77 & 0.76 & 0.47 \\
 \Addition{} & 0.6 & \PEM{} & 20825 & 3363 & 11707 & 4451 & 9025   & 0.87 & 0.70 & 0.74 & 0.79 & 0.45 \\
 \Addition{} & 0.8 & \PEM{} & 22228 & 3058 & 12975 & 4564 & 9422   & 0.89 & 0.65 & 0.73 & 0.81 & 0.41 \\ \hline
 \Brackets{} & 0.4 & \PEM{} & 3737  & 4678 & 5597  & 5635 & 6934   & 0.52 & 0.62 & 0.65 & 0.65 & 0.65 \\
 \Brackets{} & 0.6 & \PEM{} & 3931  & 4173 & 5353  & 5198 & 6138   & 0.52 & 0.58 & 0.61 & 0.62 & 0.61 \\
 \Brackets{} & 0.8 & \PEM{} & 4597  & 3848 & 3748  & 4015 & 4190   & 0.50 & 0.53 & 0.48 & 0.50 & 0.46 \\ \hline
 \Index{}    & 0.4 & \PEM{} & 4702  & 1635 & 3412  & 1965 & 3988   & 0.56 & 0.54 & 0.56 & 0.54 & 0.51 \\
 \Index{}    & 0.6 & \PEM{} & 4716  & 1820 & 3547  & 2173 & 4352   & 0.53 & 0.55 & 0.57 & 0.57 & 0.52 \\
 \Index{}    & 0.8 & \PEM{} & 3960  & 2112 & 3607  & 2566 & 4321   & 0.41 & 0.57 & 0.61 & 0.63 & 0.52 \\ \hline
 \Parity{}   & 0.4 & \PEM{} & 8932  & 5192 & 4695  & 1374 & 7607   & 0.58 & 0.73 & 0.45 & 0.44 & 0.64 \\
 \Parity{}   & 0.6 & \PEM{} & 8278  & 4890 & 6477  & 1443 & 7660   & 0.45 & 0.70 & 0.49 & 0.44 & 0.64 \\
 \Parity{}   & 0.8 & \PEM{} & 7181  & 5550 & 6964  & 1621 & 7462   & 0.41 & 0.72 & 0.54 & 0.47 & 0.62 \\ \hline
\end{tabular}
    \caption{\textbf{Efficiency parameters.} The inferred prefactors $A_{N,\tau}$ and exponents $\alpha_{N,\tau}$ of the power law linking instance size $n$ and average output length $\bar{\ell}_{N,n,\tau}$ for a correctly solved instance indicate sublinear scaling of $\bar{\ell}_{N,n,\tau}$ with $n$ across tasks and temperatures irrespective of the statistical model. The prefactors $A_{N,\tau}$ differ by task but less by temperature.}
    \label{tab:efficiency-coefficients}
\end{table}

\section{Model Comparison}\label{app:model-comparison}

Table~\ref{tab:model-comparison} summarizes the results on model comparison using leave-one-out expected log predictive density ($\elpd{}$). Higher values of this quantity indicate a better predictive performance on held-out data.

 When assessing statistical models of capability scaling, we find that the \FAM{} is preferred over the \VAM{} on the \Addition{} and \Index{} tasks for all temperatures, while the \VAM{} is favored over the \FAM{} on the \Brackets{} and \Parity{} tasks, again, for all temperatures.

 Turning to models of efficiency scaling, we find overall much smaller differences in $\elpd{}$ between the \PM{} and the \PEM{}. On the \Addition{} task the \PEM{} is the model of choice at all temperatures. On the \Brackets{} task the \PM{} is slightly preferred across temperatures. Looking at the \Index{} task, the \PM{} is favored at temperature $\tau \in \{0.4,\,0.6\}$ but narrowly disfavored at temperature $\tau = 0.8$. Finally, for the \Parity{} task the \PM{} appears to be the better statistical model across temperatures.

\begin{table}[tb]
\centering
\begin{tabular}{|l|c||c|c||c|c|}
\hline
task & $\tau$ & $\elpd{\FAM{}}$  & $\elpd{\VAM{}}$ & $\elpd{PM}$ & $\elpd{PEM}$ \\ \hhline{|=|=||=|=||=|=|}
 \Addition{} & 0.4 & \textbf{-201} & -217 &  -26.1  & \textbf{-25.3} \\
 \Addition{} & 0.6 & \textbf{-199} & -212 &  -27.6  & \textbf{-26.8}  \\
 \Addition{} & 0.8 & \textbf{-212} & -224 &  -19.0  & \textbf{-17.7} \\ \hline
 \Brackets{} & 0.4 & -301 & \textbf{-184} &  \textbf{-42.2}  & -42.3 \\
 \Brackets{} & 0.6 & -319 & \textbf{-187} &  \textbf{-28.0}  & -28.5 \\
 \Brackets{} & 0.8 & -270 & \textbf{-198} &  \textbf{-29.0}  & -30.3 \\ \hline
 \Index{}    & 0.4 & \textbf{-240} & -248 &  \textbf{-14.7}  & -16.8 \\
 \Index{}    & 0.6 & \textbf{-230} & -233 &  \textbf{-24.3}  & -26.0 \\
 \Index{}    & 0.8 & \textbf{-214} & -222 &  -25.6  & \textbf{-25.3} \\ \hline
 \Parity{}   & 0.4 & -380 & \textbf{-213} &  \textbf{-42.4}   & -42.5 \\
 \Parity{}   & 0.6 & -336 & \textbf{-209} &  \textbf{-40.2}   & -40.3 \\
 \Parity{}   & 0.8 & -343 & \textbf{-261} &  \textbf{-17.1}   & -17.2 \\ \hline
\end{tabular}
\caption{\textbf{Model comparison.} Comparing the different statistical models for capability scaling (\FAM{}, \VAM{}) and efficiency scaling (\PM{}, \PEM{}) for different tasks (\Addition{}, \Brackets{}, \Index{}, \Parity{}) and temperatures $\tau \in \{0.4, 0.6, 0.8\}$ shows that different statistical models are preferable in different settings. We indicate the preferred model, i.e., the one with higher expected log predictive density ($\elpd{}$, higher values are better), in bold.}
\label{tab:model-comparison}
\end{table}

\section{Brackets}\label{app:brackets}

\begin{figure}
    \centering
    \includegraphics[width=0.99\linewidth]{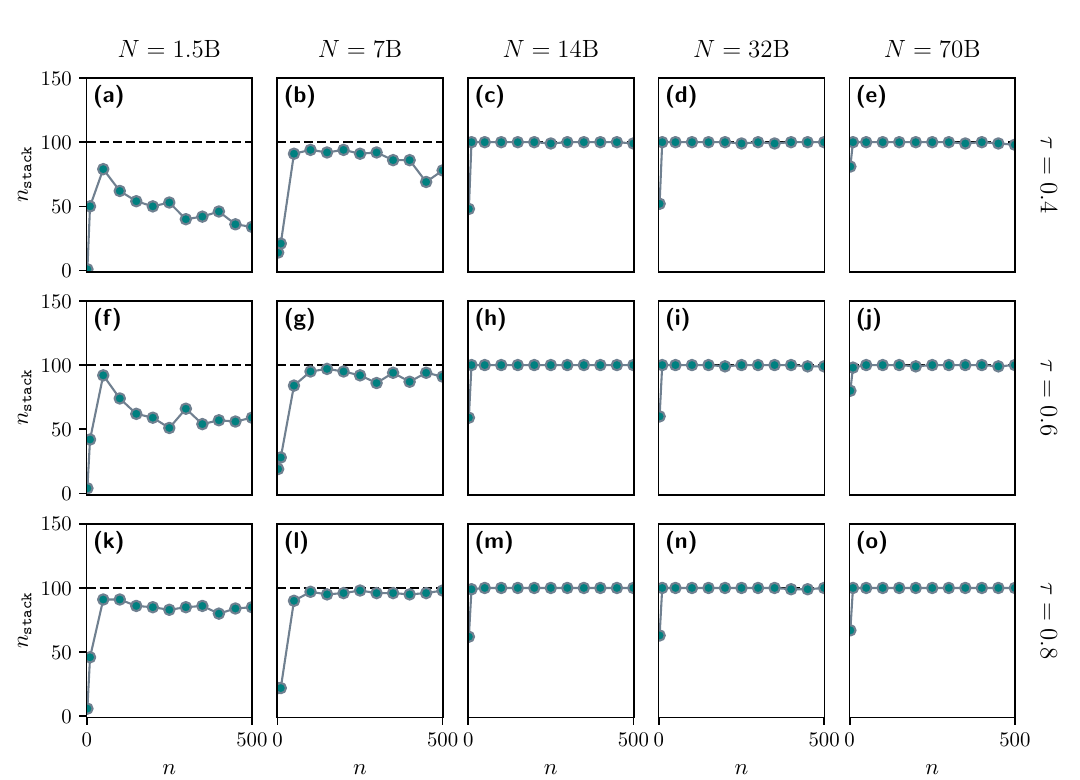}
    \caption{\textbf{Shortcut check.} The number $n_{\mathtt{stack}}$ of answers to instances of the \Brackets{} task that contain either \texttt{stack} or \texttt{Stack} as a function of the instance size $n$ for different temperatures $\tau$ (rows) and model sizes $N$ (columns) indicates that this term is common across experimental conditions, especially for models with $N\geq 7\B$ parameters. This is consistent with models using a stack-based algorithm rather than a length-based shortcut.}
    \label{fig:shortcut}
\end{figure}

Here, we provide evidence that models are not exploiting the potential shortcut in the \Brackets{} task based on the length difference of correct and incorrect instances. Figure~\ref{fig:shortcut} shows, across model sizes $N$ and temperatures $\tau$ (Fig.~\ref{fig:shortcut}\textbf{(a)}-\textbf{(e)} $\tau=0.4$, Fig.~\ref{fig:shortcut}\textbf{(f)}-\textbf{(j)} $\tau=0.6$, Fig.~\ref{fig:shortcut}\textbf{(k)}-\textbf{(o)} $\tau=0.8$), the number $n_{\mathtt{stack}}$ of answers to instances (out of a total of $R=100$ instances) of the \Brackets{} task that contain the term \texttt{stack} or \texttt{Stack}. We see that the term appears often in all combinations of model size and temperature, and is present in almost all answers in larger models, irrespective of temperature. This makes it seem plausible that models rely on stack-based algorithms instead of a length-based shortcut to solve the \Brackets{} task. Answers to individual instances that we inspected corroborate this. 

\end{document}